\documentclass{article} 
\usepackage{iclr2024_conference,times}

\usepackage{amsmath,amsfonts,bm}

\def\eqref#1{equation~\ref{#1}}

\def\1{\bm{1}}

\def\vt{{\bm{t}}}

\def\vy{{\bm{y}}}

\def\mD{{\bm{D}}}

\def\mX{{\bm{X}}}

\DeclareMathAlphabet{\mathsfit}{\encodingdefault}{\sfdefault}{m}{sl}
\SetMathAlphabet{\mathsfit}{bold}{\encodingdefault}{\sfdefault}{bx}{n}

\usepackage{hyperref}
\usepackage{url}
\usepackage{times}
\usepackage{latexsym}
\usepackage[T1]{fontenc}
\usepackage[utf8]{inputenc}
\usepackage{microtype}
\usepackage{inconsolata}
\usepackage{alltt}
\usepackage{xcolor}
\usepackage{multirow}
\usepackage[most]{tcolorbox}
\usepackage{tikz}
\usepackage{multicol,multirow}
\usepackage{color,colortbl}
\usetikzlibrary{shapes,calc,positioning}
\usepackage{cleveref}
\usepackage{stfloats}
\usepackage{booktabs}
\usepackage{xspace}
\usepackage{listings}
\usepackage{makecell}
\usepackage{colortbl}
\usepackage{wrapfig}
\usepackage{caption}
\usepackage{subcaption}
\definecolor{GREEN}{RGB}{187, 255, 185}
\definecolor{RED}{RGB}{255,200,184}

\global
\tcbset{
  aibox/.style={
    width=\textwidth,
    top=10pt,
    colback=white,
    colframe=black,
    colbacktitle=black,
    enhanced,
    center,
    attach boxed title to top left={yshift=-0.1in,xshift=0.15in},
    boxed title style={boxrule=0pt,colframe=white,},
  }
}
\newtcolorbox{AIbox}[2][]{aibox,title=#2,#1}
\tcbset{
  aiboxsmall/.style={
    width=0.48\textwidth,
    top=10pt,
    colback=white,
    colframe=black,
    colbacktitle=black,
    enhanced,
    center,
    attach boxed title to top left={yshift=-0.1in,xshift=0.15in},
    boxed title style={boxrule=0pt,colframe=white,},
  }
}
\newtcolorbox{AIboxSmall}[2][]{aiboxsmall,title=#2,#1}
\definecolor{aigold}{RGB}{244,210, 1} 
\definecolor{aired}{RGB}{255,180,181}
\newlength\savewidth\newcommand\shline{\noalign{\global\savewidth\arrayrulewidth
  \global\arrayrulewidth 1pt}\hline\noalign{\global\arrayrulewidth\savewidth}}
\newcommand{\tablestyle}[2]{\setlength{\tabcolsep}{#1}\renewcommand{\arraystretch}{#2}\centering\footnotesize}
\definecolor{defaultcolor}{gray}{0.9}

\newcommand{\shortname}{\texttt{UniNER}\xspace}
\newcommand{\longname}{\texttt{UniversalNER}\xspace}

\newcommand{\eat}[1]{\ignorespaces}

\title{\longname{}: targeted distillation from \\large language models for open named \\entity recognition}

\author{Wenxuan Zhou$^{1}$\thanks{~~Equal contributions.}~\,, Sheng Zhang$^{2*}$, Yu Gu$^{2}$, Muhao Chen$^{1,3}$, Hoifung Poon$^{2}$\\
$^{1}$University of Southern California~~~$^{2}$Microsoft Research~~~$^{3}$University of California, Davis\\ 
$^{1}${\{zhouwenx,muhaoche\}@usc.edu}\; $^{2}${{\{shezhan,yugu1,hoifung\}@microsoft.com}}\\
~\\
}

\iclrfinalcopy 
\begin{document}

\maketitle

\begin{abstract}
Large language models (LLMs) have demonstrated remarkable generalizability, such as understanding arbitrary entities and relations. 
Instruction tuning has proven effective for distilling LLMs into more cost-efficient models such as Alpaca and Vicuna. Yet such student models still trail the original LLMs by large margins in downstream applications. 
In this paper, we explore {\it targeted distillation} with mission-focused instruction tuning to train student models that can excel in a broad application class such as open information extraction.
Using named entity recognition (NER) for case study, we show how ChatGPT can be distilled into much smaller \longname models for open NER. 
For evaluation, we assemble the largest NER benchmark to date, comprising 43 datasets across 9 diverse domains such as biomedicine, programming, social media, law, finance.
Without using any direct supervision, \longname attains remarkable NER accuracy across tens of thousands of entity types, outperforming general instruction-tuned models such as Alpaca and Vicuna by over 30 absolute F1 points in average. 
With a tiny fraction of parameters, \longname not only acquires ChatGPT's capability in recognizing arbitrary entity types, but also outperforms its NER accuracy by 7-9 absolute F1 points in average. 
Remarkably, \longname even outperforms by a large margin state-of-the-art multi-task instruction-tuned systems such as InstructUIE, which uses supervised NER examples.
We also conduct thorough ablation studies to assess the impact of various components in our distillation approach. 
We release the distillation recipe, data, and \longname models to facilitate future research on targeted distillation.\footnote{Project page: \url{https://universal-ner.github.io/}}
\end{abstract}

\section{Introduction}

Large language models (LLMs) such as ChatGPT~\citep{ouyang2022training,openai2023gpt4} have demonstrated remarkable generalization capabilities, but they generally require prohibitive cost in training and inference. Moreover, in mission-critical applications such as biomedicine, white-box access to model weights and inference probabilities are often important for explainability and trust. Consequently, instruction-tuning has become a popular approach for distilling LLMs into more cost-efficient and transparent student models. 
Such student models, as exemplified by Alpaca~\citep{alpaca} and Vicuna~\citep{vicuna-2023}, have demonstrated compelling capabilities in imitating ChatGPT. However, upon close inspection, they still trail the teacher LLM by a large margin, especially in targeted downstream applications~\citep{gudibande2023false}. Bounded by limited compute, it is unsurprising that generic distillation can only produce a shallow approximation of the original LLM across all possible applications.

In this paper, we instead explore {\it targeted distillation} where we train student models using mission-focused instruction tuning for a broad application class such as open information extraction~\citep{etzioni2008open}.
We show that this can maximally replicate LLM's capabilities for the given application class, while preserving its generalizability across semantic types and domains. 
We choose named entity recognition (NER) for our case study, as it is one of the most fundamental tasks in natural language processing~\citep{wu2017clinical,perera2020named}.
Recent studies~\citep{wei2023zero,li2023evaluating} show that when there are abundant annotated examples for an entity type, LLMs still fall behind the state-of-the-art supervised system for that entity type. 
However, for the vast majority of entity types, there is little annotated data. New entity types constantly emerge, and it is expensive and time-consuming to generate annotated examples, especially in high-value domains such as biomedicine where specialized expertise is required for annotation. 
Trained on pre-specified entity types and domains, supervised NER models also exhibit limited generalizability for new domains and entity types.

We present a general recipe for targeted distillation from LLMs and demonstrate that for open-domain NER. 
We show how to use ChatGPT to generate instruction-tuning data for NER from broad-coverage unlabeled web text, and conduct instruction-tuning on LLaMA~\citep{touvron2023llama} to distill the \longname models (\shortname in short).

To facilitate a thorough evaluation, we assemble the largest and most diverse NER benchmark to date (\longname benchmark), comprising 43 datasets across 9 domains such as biomedicine, programming, social media, law, finance.
On zero-shot NER, LLaMA and Alpaca perform poorly on this benchmark (close to zero F1). Vicuna performs much better by comparison, but still trails ChatGPT by over 20 absolute points in average F1. By contrast, \longname attains state-of-the-art NER accuracy across tens of thousands of entity types in the \longname benchmark, outperforming Vicuna by over 30 absolute points in average F1. 
With a tiny fraction of parameters, \longname not only replicates ChatGPT's capability in recognizing arbitrary entity types, but also outperforms its NER accuracy by 7-9 absolute points in average F1. 
Remarkably, \longname even outperforms by a large margin state-of-the-art multi-task instruction-tuned systems such as InstructUIE~\citep{wang2023instructuie}, which uses supervised NER examples.
We also conduct thorough ablation studies to assess the impact of various distillation components, such as the instruction prompts and negative sampling. 

\section{Related Work}

\paragraph{Knowledge distillation.}
While LLMs such as ChatGPT achieve promising results, these models are often black-box and have high computational costs.
To address these issues, distilling the task capabilities of LLMs into smaller, more manageable models has emerged as a promising direction.
Knowledge distillation~\citep{hinton2015distilling} often revolves around the transfer of knowledge from larger, more complex models to their smaller counterparts.
Recent work~\citep{alpaca,vicuna-2023,peng2023instruction} seeks to distill the general abilities of LLMs with the objective of matching, if not surpassing, the performance of the original LLMs.
Particularly, Alpaca~\citep{alpaca} automates the generation of instructions~\citep{wang2022self} and distills the knowledge from a teacher LLM.
Vicuna~\citep{vicuna-2023} adopts the ShareGPT data, which are comprised of real conversations with ChatGPT conducted by users, thereby providing a more authentic context for distillation.
Another line of work~\citep{smith2022language,jung2023impossible,hsieh-etal-2023-distilling,gu2023distilling} focuses on distilling task-level abilities from LLMs.
Particularly, \citet{jung2023impossible} propose an efficient method to distill an order of magnitude smaller model that outperforms GPT-3 on specialized tasks summarization and paraphrasing in certain domains.
\citet{hsieh-etal-2022-converting} propose to distill LLMs' reasoning abilities into smaller models by chain-of-the-thought distillation.
However, these studies perform distillation either on certain datasets or domains, while our work focuses on a more general formulation that can be applied to diverse domains.

\paragraph{Instruction tuning.} 
As an effective method to adapt LMs to perform a variety of tasks, instruction tuning has attracted an increasing number of community efforts: FLAN~\citep{chung2022scaling}, T0~\citep{sanh2021multitask}, and Tk-Instruct~\citep{wang-etal-2022-super} convert a large set of existing supervised learning datasets into instruction-following format, and then fine-tune encoder-decoder models, showing strong zero-shot and few-shot performance on NLP benchmarks. \citet{ouyang2022training} crowd-source high-quality instruction data and fine-tune GPT-3 into InstructGPT, enhancing its ability to understand user intention and follow instructions.
Recent advancements~\citep{alpaca,vicuna-2023,peng2023instruction} have also led to smaller models that exhibit task-following capabilities, after being fine-tuned on instruction data generated by LLMs, such as ChatGPT or GPT4.
However, these smaller models often struggle to generate high-quality responses for a diverse range of tasks~\citep{wang2023far}.
A closer examination on targeted benchmarks reveals a substantial gap between these models to ChatGPT~\citep{gudibande2023false}. 
Our proposed method, in contrast, focuses on tuning models to excel at a specific type of tasks. 
The diversity in our instructing-tuning method comes from task labels (e.g., relation types for relation extraction, entity types for NER), rather than instructions.
By focusing on task-level capabilities and using NER as a case study, we demonstrate that it is possible to devise a tuning recipe that not only closes the performance gap but also surpasses ChatGPT.
\citet{wang2023instructuie} also explore instruction-tuning for information extraction tasks. However, their method relies solely on supervised datasets and yields subpar performance when compared to ChatGPT.

\section{Mission-Focused Instruction Tuning}

Instruction tuning~\citep{ouyang2022training,wei2022finetuned} is a method through which pretrained autoregressive language models 
are finetuned to follow natural language instructions and generate responses.
Existing work focuses on tuning models to do diverse tasks~\citep{alpaca,vicuna-2023}. In contrast, we introduce a general recipe for mission-focused instruction tuning, where the pretrained model is tuned for a broad application class such as open information extraction.

In this paper, we conduct a case study on the NER task, as it is one of the fundamental tasks for knowledge extraction from text.
The objective is to learn a model $f: (\mathcal{X} \times \mathcal{T})\rightarrow \mathcal{Y}$, where $\mathcal{X}$ represents the set of inputs, $\mathcal{T}$ denotes a predefined set of entity types, and $\mathcal{Y}$ represents the set of entities of a specific type in the given input.

\subsection{Data Construction}
A typical instruction-tuning example is made of three parts, including \texttt{instruction}, \texttt{input}, and \texttt{output}, where the diversity of \texttt{instruction} 
causes the models to follow
a wide range of task instructions.
However, for \emph{mission-focused} instruction tuning, our goal is to tune the model to maximally generalize across semantic types and domains for the targeted application class.
Therefore, we focus on increasing the diversity of \texttt{input} rather than \texttt{instruction}.

\begin{wrapfigure}[16]{r}{0.5\textwidth}
    \vspace{-10pt}
    \begin{AIboxSmall}{\footnotesize Data Construction Prompt}
    \footnotesize
    \textbf{System Message:} You are a helpful information extraction system.
    
    \textbf{Prompt:} Given a passage, your task is to extract all entities and identify their entity types. The output should be in a list of tuples of the following format: [("entity 1", "type of entity 1"), ... ].\\

Passage: \{input\_passage\}
    \end{AIboxSmall}
    \caption{Data construction prompt for generating entity mentions and their types for a given passage.}
    \label{fig:prompt_ent_type_gen}
\end{wrapfigure}

While earlier work~\citep{jung2023impossible} employs language models to generate inputs, these models typically assume that the domains of test data are known and prompt LMs to generate data for each domain.
This method falls short when applied to distillation for a broad application class, where the distribution of test data is unknown.
Consequently, it is challenging to generate inputs from LMs that provide wide coverage of the test domains.

To address this limitation, we propose an alternative: directly sampling inputs from a large corpus across diverse domains, and then using an LLM to generate outputs.
In this paper, we sample inputs from the Pile corpus~\citep{gao2020pile}, which compiles 22 distinct English sub-datasets.
We chunk the articles in Pile to passages of a max length of 256 tokens and randomly sample 50K passages as the inputs.
Subsequently, we use ChatGPT (\texttt{gpt-3.5-turbo-0301}) to generate entity mentions and their associated types based on the sampled passages.
To ensure stability, we set the generation temperature to 0.
The specific prompt for constructing the data is shown in Fig.~\ref{fig:prompt_ent_type_gen}.
In this prompt, we do not specify the set of entity types of interest, allowing the LLM to generate outputs encompassing a broad coverage of entity types.

\begin{table}[!t]
    \centering
    \small
    \begin{tabular}{c|p{0.8\linewidth}}
        \shline
         Frequency&  Entity types\\
         \shline
         \makecell[tc]{Top 1\% \\ (74\%)} & person, organization, location, date, concept, product, event, technology, group, medical condition, ...\\
         \shline
         \makecell[tc]{1\%-10\% \\ (19\%)}& characteristic, research, county, module, unit, feature, cell, package, anatomical structure, equipment, ... \\
         \shline
         \makecell[tc]{10\%-100\% \\ (7\%)}& attribute value, pokemon, immune response, physiology, animals, cell feature, FAC, input device, ward, broadcast, ... \\
         \shline
    \end{tabular}
    \caption{Examples of entities across different frequency ranges - top 1\%, 1-10\%, and 10-100\%, along with the percentage of total frequencies for each range.}
    \label{tab:entity_types}
\end{table}

\smallskip
\noindent\textbf{Data statistics.} After filtering out unparseable outputs and inappropriate entities, including non-English entities and those classified under 'ELSE' categories, such as None, NA, MISC, and ELSE, our dataset comprises 45,889 input-output pairs, encompassing 240,725 entities and 13,020 distinct entity types.
We divide the entity types according to frequency and show the top 10 entity types in each range in Tab.~\ref{tab:entity_types}.
The distribution of these entity types exhibits a heavy tail, where the top 1\% of entities account for 74\% of total frequencies.
We find that the generated data contain entity types from various domains, ranging from the general domain (e.g., \textsc{person}) to the clinical domain (e.g., \textsc{medical condition}).
Moreover, we observe variations in granularity among the entity types.
E.g., \textsc{county} is the subset of \textsc{location}, and \textsc{input device} is a subset of \textsc{product}.
These data characteristics offer extensive coverage of entity types, making them suitable for distilling capabilities from LLMs across various domains.

\smallskip
\noindent\textbf{Definition-based data construction.} Besides entity types, we also prompt ChatGPT to generate entity mentions and define their types using short sentences.
To do so, we simply change the prompt in Fig.~\ref{fig:prompt_ent_type_gen} from ``extract all entities and identify their entity types'' to ``extract all entities and concepts, and \emph{define their type using a short sentence}''.
This method generates a much more diverse set of 353,092 entity types and leads to a tuned model that is less sensitive to entity type paraphrasing (Section~\ref{ssec:case_study}), but performs worse on standard NER benchmarks (Section~\ref{ssec:results_distill}).

    


\subsection{Instruction Tuning}

\begin{wrapfigure}[20]{r}{0.5\textwidth}
    \vspace{-20pt}
    \begin{AIboxSmall}{\footnotesize Conversation-style Instruct Tuning Template}
    \footnotesize
    A virtual assistant answers questions from a user based on the provided text.\\
    \texttt{User:} Text: $\mX_\text{passage}$\\
    \texttt{Assistant:} I've read this text.\\
    \texttt{User:} What describes $\vt_1$ in the text?\\
    \texttt{Assistant:} \fcolorbox{aigold}{aigold}{$\vy_1$}\\
    ...\\
    \texttt{User:} What describes $\vt_T$ in the text?\\
    \texttt{Assistant:} \fcolorbox{aigold}{aigold}{$\vy_T$}
    \end{AIboxSmall}
    \caption{The conversation-style template that converts a passage with NER annotations into a conversation, where $\mX_\text{passage}$ is the input passage, $[\vt_1,...,\vt_T]$ are entity types to consider, and $\vy_i$ is a list of entity mentions that are $\vt_i$. The conversation is used to tune language models. Only the highlighted parts are used to compute the loss.}
    \label{fig:input_in_conv}
\end{wrapfigure}

After obtaining the data, we apply instruction tuning to smaller models to distill for a broad application class, e.g., diverse entity types in NER.
Our template, as shown in Fig.~\ref{fig:input_in_conv}, adopts a conversation-style tuning format.
In this approach, the language model is presented with a passage $\mX_\text{passage}$ as input.
Then, for each entity type $\vt_i$ that appears in the output, we transform it into a natural language query ``\textit{What describes $\vt_i$?}''
Subsequently, we tune the LM to generate a structured output $\vy_i$ in the form of a JSON list containing all entities of $\vt_i$ in the passage.
We consider $\vy_1,...,\vy_T$ as gold tokens and apply a language modeling objective on these tokens.
Our preliminary experiments show that conversation-style tuning is better than traditional NER-style tuning adopted by \citet{wang2023instructuie,sun2023pushing}.


Besides one entity type per query, we also consider combining all entity types in a single query, requiring the model to output all entities in a single response.
Detailed results and discussions can be found in Section~\ref{ssec:results_distill}.

\smallskip
\noindent\textbf{Negative sampling.}
Our data construction process follows an open-world assumption where we allow the model to generate entity types that have appeared in the passage.
However, the generated data do not account for entity types that are not mentioned in the passage, i.e., negative entity types.
As a result, it is challenging for us to apply a model trained on this data to a closed-world setting, where one may ask for entity types that do not exist in the passage.
To address this potential mismatch, we sample negative entity types from the collection of all entity types that do not appear in the passage as queries and set the expected outputs as empty JSON lists.
The sampling of negative entity types is done with a probability proportional to the frequency of entity types in the entire dataset.
This approach greatly improves the instruction tuning results, as shown in Section~\ref{ssec:ablation_study}.

\smallskip
\noindent\textbf{Supervised finetuning.} When we have additional human annotations, model performance can be further improved with supervised data.
However, a significant challenge arises when training with multiple datasets, as there might be discrepancies in label definitions among these datasets, resulting in label conflicts.
For instance, some datasets like ACE~\citep{walker2006ace} consider personal pronouns (e.g., she, he) as \textsc{person}, while other datasets like multiNERD~\citep{tedeschi-navigli-2022-multinerd} do not include pronouns.

\begin{wrapfigure}[13]{r}{0.5\textwidth}
    \vspace{-20pt}
    \begin{AIboxSmall}{\footnotesize Dataset-specific Instruct Tuning Template}
    \footnotesize
    A virtual assistant answers questions from a user based on the provided text.\\
    \texttt{User:} \fcolorbox{aired}{aired}{Dataset: $\mD$} \textbackslash n Text: $\mX_\text{passage}$\\
    \texttt{Assistant:} I've read this text.\\
    \texttt{User:} What describes $\vt_1$ in the text?\\
    \texttt{Assistant:} \fcolorbox{aigold}{aigold}{$\vy_1$}\\
    ...\\
    \texttt{User:} What describes $\vt_T$ in the text?\\
    \texttt{Assistant:} \fcolorbox{aigold}{aigold}{$\vy_T$}
    \end{AIboxSmall}
    \caption{The dataset-specific instruction tuning template. We add the dataset name $\mD$ (colored in \fcolorbox{aired}{aired}{red}) as part of the input to resolve conflicts in label definitions.}
    \label{fig:supervised_input_in_conv}
\end{wrapfigure}

To address this issue, we propose to use dataset-specific instruction tuning templates to harmonize the discrepancies in label definitions, as illustrated in Fig.~\ref{fig:supervised_input_in_conv}.
Specifically, we augment the input with an additional field denoting the dataset name $\mD$.
By doing so, the model can learn the dataset-specific semantics of labels.
During inference, we use the respective dataset name in the prompt for the supervised setting, whereas we omit the dataset field from the prompt in the zero-shot setting.

\section{Universal NER Benchmark}
To conduct a comprehensive evaluation of NER models across diverse domains and entity types, we collect the largest NER benchmark to date.
This benchmark encompasses 43 NER datasets across 9 domains, including general, biomedical, clinical, STEM, programming, social media, law, finance, and transportation domains.
An overview of data distribution is shown in Fig.~\ref{fig:data_dist}.
Detailed dataset statistics are available in Appendix Tab.~\ref{tab:full_data_statistics}.

\begin{wrapfigure}[22]{r}{0.5\textwidth}
    \vspace{-15pt}
    \centering
    \includegraphics[width=1.0\linewidth]{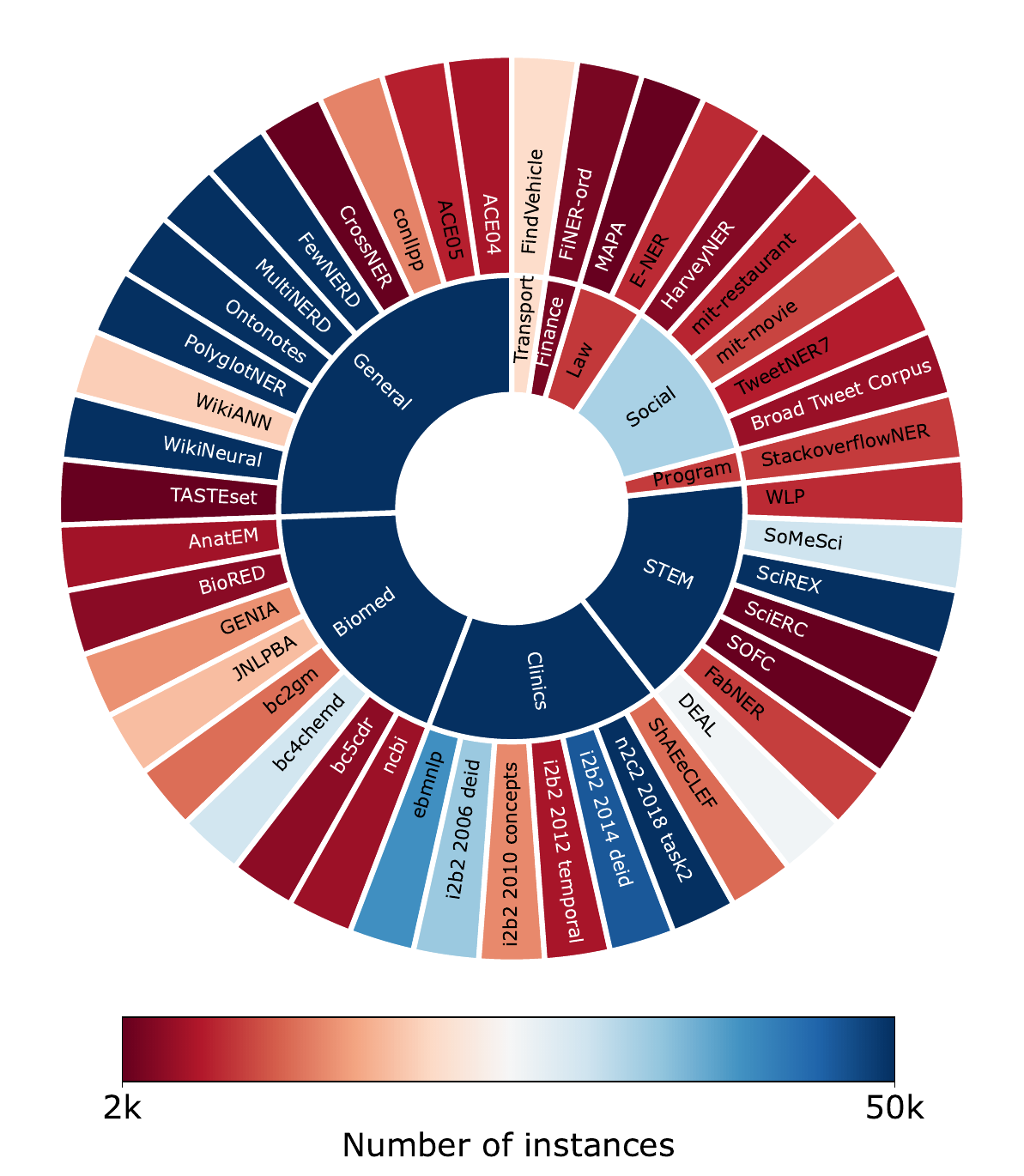}
    \caption{Distribution of UniNER benchmark.}
    \label{fig:data_dist}
\end{wrapfigure}

\smallskip
\noindent\textbf{Dataset processing.} To make the entity types semantically meaningful to LLMs, we conduct a manual inspection of the labels and convert the original labels into natural language formats.
For instance, we replace \textsc{per} with \textsc{person}.
While we try to collect a broad coverage of NER datasets, we do not use all entity types.
This is because some entity types (e.g., \textsc{Else}) are not coming from consistent sources across the different datasets.
Their annotations often come from different ontologies for different purposes.
The choices of entity types and their annotation guidelines are not optimized for holistic or comprehensive assessments, which renders them suboptimal for use as a ``ground truth'' to evaluate a universal NER model.
Therefore, we remove those labels from the datasets.
In addition, some datasets are at the document level and contain very long contexts, which might exceed the input length limit of models.
Therefore, we split all instances in document-level datasets into sentence-level 
ones.

\section{Experiments}
This section presents experimental evaluations of \longname.
We start by outlining experimental settings (Section~\ref{ssec:experiment_setting}), followed by presenting the results on both distillation and supervised settings (Sections~\ref{ssec:results_distill} and~\ref{ssec:results_supervised}).
Finally, we conduct analysis (Section~\ref{ssec:ablation_study}) and case study (Section~\ref{ssec:case_study}) to provide deeper insights into the model's performance.

\begin{figure*}[t]
\begin{subfigure}[t]{0.46\textwidth}
\centering
\includegraphics[width=0.99\textwidth]{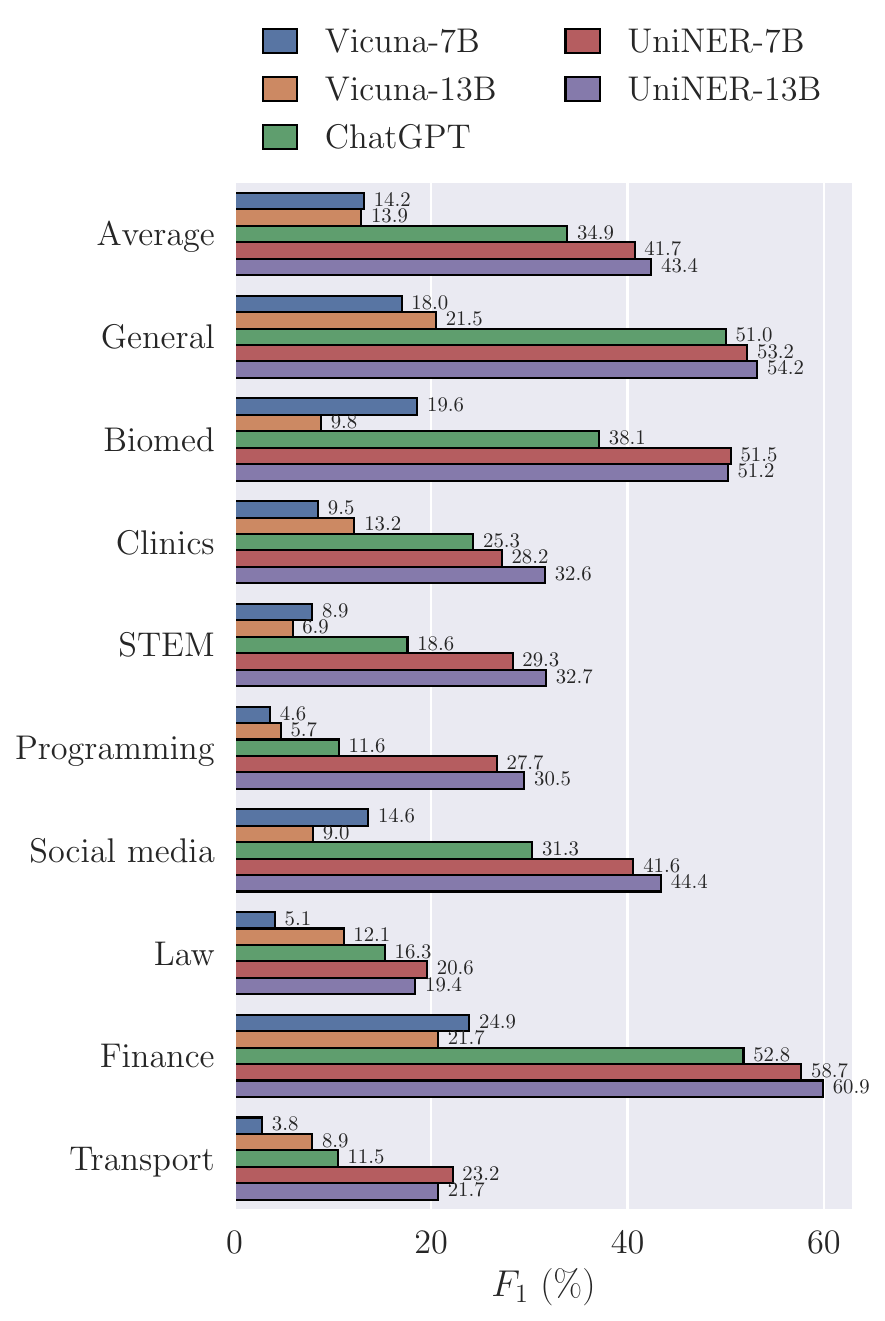}
\caption{Comparisons of zero-shot models on different domains. Our distilled models achieve better results than ChatGPT in all evaluated domains.}
\label{fig:self-supervised-bar}
\end{subfigure}
\hfill
\begin{subfigure}[t]{0.46\textwidth}
\centering
\includegraphics[width=0.97\textwidth]{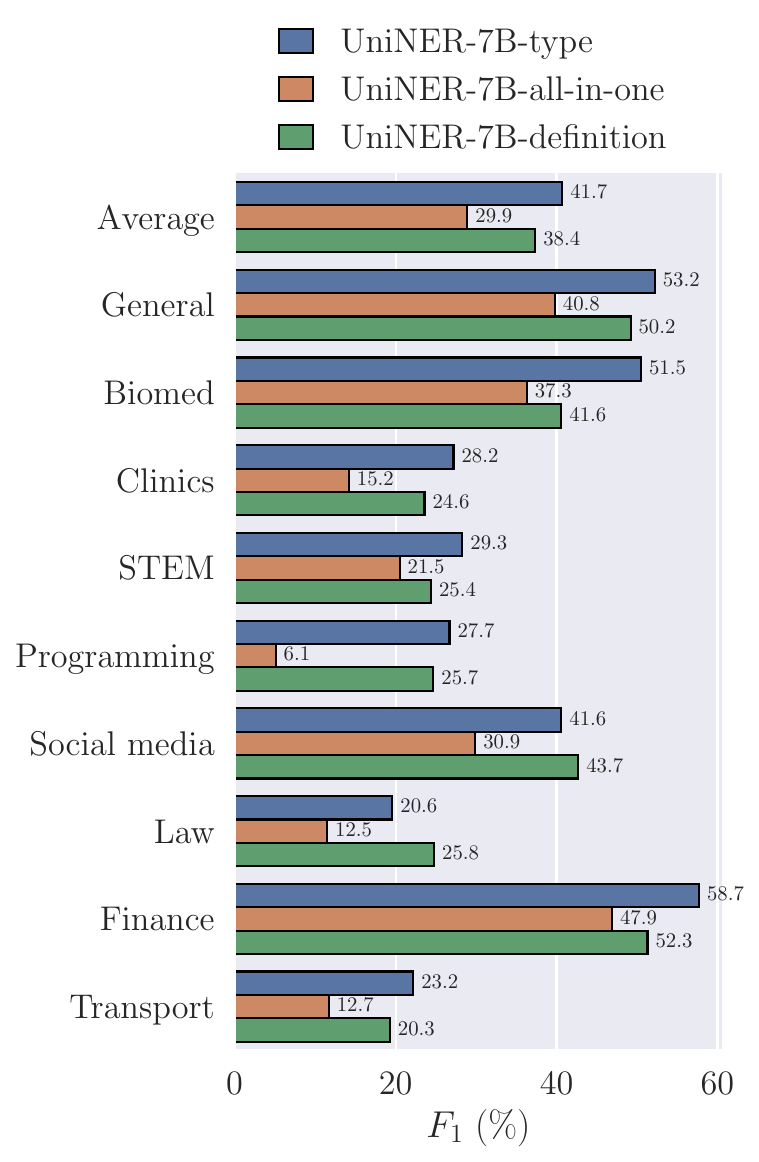}
\caption{Comparisons between \shortname-7B and two variants. \shortname-7B-definition is distilled on Pile data prompted with entity type definitions. \shortname-7B-all-in-one is tuned with the template where all entity types are asked in one query.}
\label{fig:self-supervised-ablation}
\end{subfigure}
\end{figure*}

\subsection{Experimental Settings}
\label{ssec:experiment_setting}
\smallskip
\noindent\textbf{Model configurations.}
We train models based on LLaMA\footnote{We also train models based on LLaMA 2~\citep{touvron2023llama2}. However, no significant difference is observed in our experiments.}~\citep{touvron2023llama} following the training schedule of~\citet{vicuna-2023} for a fair comparison.
Considering the large size of certain test datasets, we perform evaluation by sampling up to 200,000 passage-query pairs from each dataset.
We use strict entity-level micro-$F_1$ in evaluation, requiring both the entity type and boundary to exactly match the ground truth.

\smallskip
\noindent\textbf{Compared models.}
We compare our model (\shortname{}) against the following models: (1) \textbf{ChatGPT} (\texttt{gpt-3.5-turbo-0301}).
We use the prompting template in~\citet{ye2023comprehensive} for NER.
(2) \textbf{Vicuna}~\citep{vicuna-2023} is finetuned with ChatGPT conversations, using LLaMA as the base model.
(3) \textbf{InstructUIE}~\citep{wang2023instructuie} is a supervised model finetuned on diverse information extraction datasets, employing a unified natural language generation objective.
It adopts Flan-T5 11B~\citep{chung2022scaling} as the base model.

\subsection{Distillation}
\label{ssec:results_distill}
We first evaluate the models in a zero-shot setting.
We compare the performance of ChatGPT, Vicuna, and our model \shortname, which is distilled from ChatGPT NER annotations on Pile without human-labeled datasets in training.
Results are shown in Fig.~\ref{fig:self-supervised-bar}.\footnote{Due to limited space, we only show the average $F_1$ of all datasets and the average $F_1$ of each domain. See Appendix Fig.~\ref{fig:add-supervised-bar} for full results.}
We observe that our distilled models, namely \shortname-7B and \shortname-13B, outperform ChatGPT in terms of average $F_1$.
The average $F_1$ scores of \shortname-7B and \shortname-13B are 41.7\% and 43.4\%, respectively, compared to 34.9\% for ChatGPT.
This demonstrates that our proposed targeted distillation from diverse inputs yields models that have superior performance on a broad application class while maintaining a relatively small model size.
Additionally, \shortname-13B exhibits better performance compared to \shortname-7B, indicating that fine-tuning on larger models may lead to improved generalization.
In terms of domains, both \shortname-7B and \shortname-13B outperform ChatGPT on all domains, showing that the improvements exist across various domains.

\begin{wraptable}{r}{6cm}
\vspace{-15pt}
\centering
\tablestyle{4pt}{1.3}
\setlength{\tabcolsep}{2pt}
\begin{tabular}{c|cc|c}
\shline
& BERT- & InstructUIE & \shortname \\
Dataset & base & 11B & 7B \\
\shline
ACE05 & \textbf{87.30} & 79.94 & 86.69 \\
AnatEM & 85.82 & 88.52 & \textbf{88.65} \\
bc2gm & 80.90 & 80.69 & \textbf{82.42} \\
bc4chemd & 86.72 & 87.62 & \textbf{89.21} \\
bc5cdr & 85.28 & 89.02 & \textbf{89.34} \\
Broad Twitter & 58.61 & 80.27 & \textbf{81.25} \\
CoNLL03 & 92.40 & 91.53 & \textbf{93.30}\\
FabNER & 64.20 & 78.38 & \textbf{81.87}\\
FindVehicle & 87.13 & 87.56 & \textbf{98.30} \\
GENIA & 73.3 & 75.71 & \textbf{77.54} \\
HarveyNER & \textbf{82.26} & 74.69 & 74.21 \\
MIT Movie & 88.78 & 89.58 & \textbf{90.17} \\
MIT Restaurant & 81.02 & \textbf{82.59} &82.35 \\
MultiNERD & 91.25 & 90.26 & \textbf{93.73} \\
ncbi & 80.20 & 86.21 & \textbf{86.96} \\
OntoNotes & \textbf{91.11} & 88.64 & 89.91 \\
PolyglotNER & \textbf{75.65} & 53.31 & 65.67 \\
TweetNER7 & 56.49 & \textbf{65.95} & 65.77 \\
WikiANN & 70.60 & 64.47 & \textbf{84.91} \\
wikiNeural & 82.78 & 88.27 & \textbf{93.28} \\ \hline
Avg & 80.09 & 81.16 & \textbf{84.78} \\
\shline
\end{tabular}
\caption{$F_1$ on 20 datasets used in \citet{wang2023instructuie}. BERT-base results are from~\citet{wang2023instructuie}. InstructUIE results are from our reevaluation.}
\label{tab:supervised-eval}
\vspace{-38pt}
\end{wraptable}

We further compare different variations of \shortname, including (1) \shortname-all-in-one, where the extraction of all entity types are combined into one query and response, and (2) \shortname-definition, where queries in instruction tuning data use entity type definitions generated by ChatGPT instead of entity types.
Results are shown in Fig.~\ref{fig:self-supervised-ablation}.
We observe that both \shortname-all-in-one and \shortname-definition underperform \shortname-type by 3.3\% and 11.8\% on average, respectively.
The \shortname-definition variant's decreased performance could be due to its lower consistency with the evaluation datasets, which all adopt words or short phrases as labels instead of sentences.
The performance disparity in the \shortname-all-in-one variant can be potentially attributed to the attention distribution and task complexity.
When the model is required to handle multiple entity types within a single query, it might disperse its attention across these varied types, possibly resulting in less accurate identification for each individual type.
Conversely, by decomposing the task into several simpler ones, each focusing on one entity type at a time, the model might be better equipped to handle the complexity, thus yielding more accurate results.

\subsection{Supervised Finetuning}
\label{ssec:results_supervised}
We study whether our models can be further improved using additional human annotations.
We compare the performance of ChatGPT, Vicuna, InstructUIE~\citep{wang2023instructuie}~\footnote{Please note that the original evaluation script in InstructUIE contains a critical bug. For passages that do not contain any entities, the script adds \textsc{none} as a placeholder entity and takes it into account when calculating $F_1$. To rectify this error, we re-evaluated InstructUIE using their released checkpoint.}, and \shortname.

\smallskip
\noindent\textbf{Out-of-domain evaluation.} We first study whether supervised finetuning leads to better generalization on unseen data.
We follow InstructUIE to exclude two datasets CrossNER~\citep{liu2021crossner} and MIT~\citep{liu2013asgard} for out-of-domain evaluation, and fine-tune our model using training splits of the remaining datasets in the universal NER benchmark.
Results are shown in Tab.~\ref{tab:zero_shot_2}.
Notably, without any fine-tuning, instruction-tuned \shortname{} 7B and 13B already surpass ChatGPT, Vicuna, and the supervised fine-tuned InstructUIE-11B by a large margin.
If we train our model from scratch only using the supervised data, it achieves an average $F_1$ of 57.2\%.
Continual fine-tuning \shortname-7B using the supervised data achieves the best average $F_1$ of 60.0\%.
These findings suggest that the models' generalization can be further improved with additional human-annotated data.

\begin{table*}[!t]
\centering
\tablestyle{4pt}{1.1}
\begin{tabular}{l|ccccccc|c}
\shline
\multicolumn{1}{c|}{Model} & Movie & Restaurant & AI & Literature & Music & Politics & Science & Avg  \\ \shline
\emph{Zero-shot} & & & & & & & & \\
\quad Vicuna-7B& 6.0& 5.3& 12.8& 16.1& 17.0& 20.5& 13.0& 13.0 \\
\quad Vicuna-13B& 0.9& 0.4& 22.7& 22.7& 26.6& 27.2& 22.0& 17.5 \\
\quad ChatGPT & 5.3 & 32.8 & 52.4 & 39.8 & \textbf{66.6} & \textbf{68.5} & \textbf{67.0} & 47.5 \\
\quad \shortname-7B & 42.4& 31.7& 53.5& 59.4& 65.0 & 60.8& 61.1& 53.4 \\ 
\quad \shortname-13B & \textbf{48.7} & \textbf{36.2} & \textbf{54.2} & \textbf{60.9} & 64.5& 61.4 & 63.5 & \textbf{55.6} \\ \shline
\emph{In-domain supervised} & & & & & & & & \\
\quad InstructUIE-11B  & - & - & 48.4 & 48.8 & 54.4 & 49.9 & 49.4 & - \\
\quad \shortname-7B (sup. only) & 54.2& 16.0& 62.3& \textbf{67.4}& 69.0& 64.5& 66.9& 57.2 \\
\quad \shortname-7B (inst-tuned + sup.) & \textbf{61.2}& \textbf{35.2}& \textbf{62.9}& 64.9& \textbf{70.6}& \textbf{66.9}& \textbf{70.8}& \textbf{61.8} \\
\shline
\end{tabular}
\caption{Out-of-domain evaluation on datasets from \citet{wang2023instructuie}. ``sup. only'' denotes a variant of \shortname-7B, trained from scratch using in-domain supervised data only and evaluated on out-of-domain datasets.}
\label{tab:zero_shot_2}
\end{table*}


\smallskip
\noindent\textbf{In-domain evaluation.}
We then study the performance of \shortname in an in-domain supervised setting, where we fine-tune \shortname-7B using the same training data as InstructUIE~\citep{wang2023instructuie}.
Results are shown in Tab.~\ref{tab:supervised-eval}.
Our \shortname-7B achieves an average $F_1$ of 84.78\% on the 20 datasets, surpassing both BERT-base and InstructUIE-11B by 4.69\% and 3.62\%, respectively.
This experiment demonstrates the effectiveness of our model in the supervised setting.

\subsection{Analysis}
\label{ssec:ablation_study}

\begin{table*}[!ht]
\centering
\tablestyle{4pt}{1.3}
\begin{tabular}{l|ccccccc|c}
\shline
Strategy & Movie & Restaurant & AI & Literature & Music & Politics & Science & Avg  \\ \shline
None & 19.1& 19.1& 25.1& 39.5& 42.7& 48.9& 26.2& 31.5 \\
Uniform & 42.5& 29.0& 42.5& 53.3& 57.4& 56.8& 52.6& 47.7 \\
\rowcolor{defaultcolor} Frequency &42.4& 31.7& 53.5& 59.4& 65.0& 60.8& 61.1& 53.4 \\
\shline
\end{tabular}
\caption{Ablation study on negative sampling strategies for \shortname-7B. All models are instruction-tuned on Pile.}
\label{tab:neg-sample-ablation}
\end{table*}

\begin{wrapfigure}[23]{r}{0.5\textwidth}
\vspace{-22pt}
    \centering
    \includegraphics[width=0.48\textwidth]{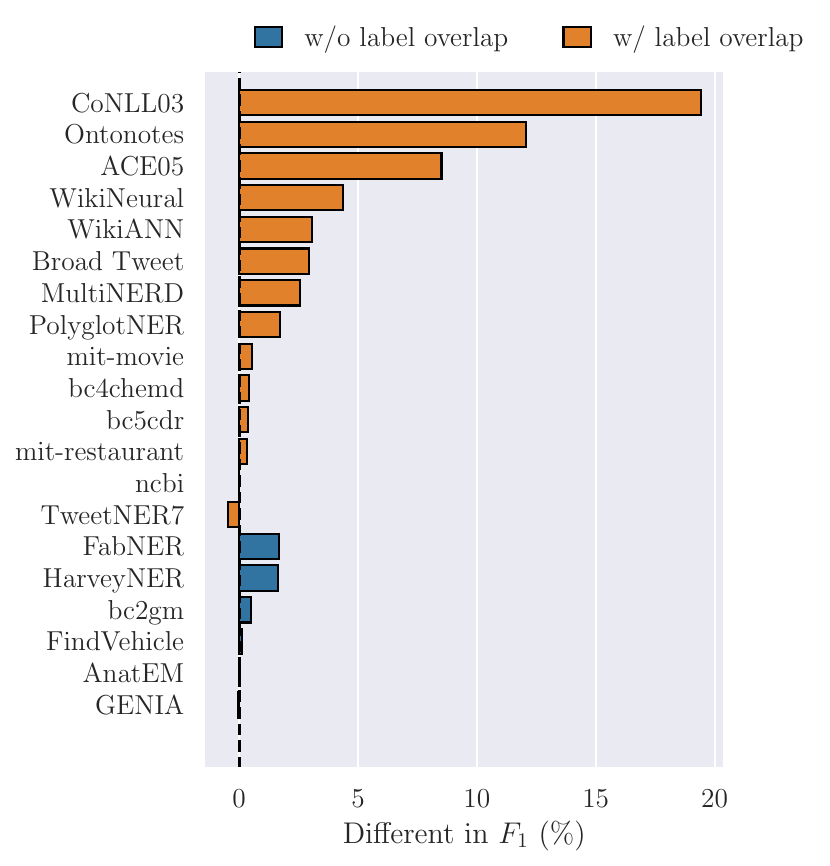}
    \caption{Different in $F_1$ between data-specific and original templates in the supervised setting. \textcolor[HTML]{ff7f0e}{Orange} and \textcolor[HTML]{1f77b4}{Blue} mark datasets with/without label overlap with other datasets, respectively.}
    \label{fig:dataset_specific_inst}
\end{wrapfigure}

\smallskip
\noindent\textbf{Negative sampling strategies.}
We experiment with different negative sampling strategies in instruction tuning, including (1) \emph{no negative sampling}, (2) \emph{uniform sampling} where entity types are randomly sampled with equal probability for each one, and (3) \emph{frequency-based sampling} where we sample entity types with probabilities proportional to their frequency in the constructed dataset.
Results are shown in Tab.~\ref{tab:neg-sample-ablation}.
Among the approaches tested, frequency-based sampling yielded the best results, outperforming no sampling and uniform sampling by 21.9\% and 5.7\%, respectively. 
These findings highlight the crucial role of negative sampling in instruction tuning, with frequency-based sampling emerging as the most effective method for enhancing model performance in our study.

\smallskip
\noindent\textbf{Dataset-specific template.}
We compare the results of our dataset-specific instruction tuning template and the original template in the supervised setting.
As shown in Fig.~\ref{fig:dataset_specific_inst},
we find that the data-specific template outperforms the original template on most datasets.
To gain deeper insights into the improvements achieved, we further divide the datasets into two categories: those with label (entity type) overlap with other datasets and those without overlap. 
Our analysis reveals that datasets with label overlap demonstrate more substantial improvements.

To explore this further, we measure $F_1$ score across all evaluation datasets and calculate the difference. Apart from the long-tail entity types that manifest a high variance in results, we identify two entity types where the dataset-specific template outperforms the original template by over 10\%: \textsc{facility} (22.0\%) and \textsc{time} (12.4\%). 
Intriguingly, both labels exhibit inconsistencies in their definitions across various datasets.
The \textsc{facility} label has been annotated on pronouns (e.g., it, which) as entities in ACE datasets but are excluded in OntoNotes.
The \textsc{time} label denotes well-defined time intervals (e.g., Christmas) in MultiNERD, but may encompass any general time expressions (e.g., 3 pm) in OntoNotes.
This finding suggests that the improvements provided by the data-specific template are particularly effective in resolving label conflicts.

\begin{table*}[!t]
\centering
\tablestyle{3pt}{1.2}
\begin{tabular}{c|lrcccccc|c}
\shline
Partial match & Model & Movie & Restaurant & AI & Literature & Music & Politics & Science & Avg  \\ \shline
\multirow{3}{*}{No}& ChatGPT & 5.3 & 32.8 & 52.4 & 39.8 & 66.6 & 68.5 & 67.0 & 47.5 \\
& \shortname-7B & 42.4& 31.7& 53.5& 59.4& 65.0& 60.8& 61.1& 53.4 \\
& \shortname-7B w/ sup & 61.2& 35.2& 62.9& 64.9& 70.6& 66.9& 70.8& 61.8 \\
\shline
\multirow{3}{*}{Yes}& ChatGPT& 5.9& 40.1& 55.7& 42.8& 70.2& 71.7& 70.1& 50.9 \\
& \shortname-7B & 46.9& 40.3& 57.7& 62.7& 62.9& 63.2& 63.3& 56.7 \\
& \shortname-7B w/ sup & 65.5& 39.4& 66.2& 67.2& 72.7& 68.9& 73.4& 64.8 \\
\shline
\end{tabular}
\caption{Allowing partial match between the prediction and the gold that has overlap increases the results.
When it is allowed, any partial match is regarded as half correct (counted as 0.5 in true positive) when computing $F_1$.
}
\label{tab:loose-$F_1$}
\end{table*}

\smallskip
\noindent\textbf{Evaluation with partial match.}
While using strict $F_1$ as an evaluation metric, we notice that it may underestimate the zero-shot learning capabilities of NER models.
In particular, strict $F_1$ penalizes slight misalignments in the boundaries of the extracted entities, which may not necessarily indicate an incorrect understanding of the text. 
For instance, given the sentence \textit{any asian cuisine around} and the entity type \textsc{cuisine}, \shortname extracts \textit{asian cuisine} as the named entity, while the ground truth only labels \textit{asian} as the correct entity.
However, the model's prediction can still be viewed as correct, even though it is deemed incorrect by strict $F_1$.
To better estimate the zero-shot abilities, we also consider partial match~\citep{segura2013semeval} in evaluation.
In this context, a prediction that exhibits word overlap with the ground truth is regarded as half correct (counted as 0.5 in true positive) when computing $F_1$.
Results are shown in Tab.~\ref{tab:loose-$F_1$}.
We find that allowing partial match consistently improves the results.
Besides, our models is still the best-performing model on average.

\subsection{Case Study}
\label{ssec:case_study}

\smallskip
\noindent \textbf{Sensitivity to entity type paraphrasing.}
One type of entity can be expressed in multiple ways, so it is essential for our model to give consistent predictions given entity types with similar meanings.
An example of sensitivity analysis is present in Fig.~\ref{fig:sensitivity}.
We observe that \shortname-7B-type 
sometimes fails to recognize entities with similar semantic meanings.
On the other hand, \shortname-7B-definition, despite performing worse on our Universal NER benchmark, exhibits robustness to entity type paraphrasing.
It demonstrates that although using definitions may result in lower performance on standard NER benchmarks, it could yield improved performance for less populous entity types.

\smallskip
\noindent \textbf{Recognition of diverse entity types.}
We present an example in Fig.~\ref{fig:code-ex} showcasing the capabilities of \shortname in recognizing various entities.
Particularly, we focus on a novel domain of code and assess \shortname's ability to extract diverse types of entities within the code.
Despite minor mistakes (e.g., \texttt{from\_pretrained} is not identified as a method), this case study effectively demonstrates our model's capacity to capture entities of various types.

\section{Conclusion}
We present a targeted distillation approach with mission-focused instruction tuning. Using NER as a case study, we train smaller and more efficient models for open-domain NER. The proposed method successfully distills ChatGPT into a smaller model \longname{}, achieving remarkable NER accuracy across a wide range of domains and entity types without direct supervision. These models not only retain ChatGPT's capabilities but also surpass it and other state-of-the-art systems in NER performance.

\section*{Acknowledgement}

Wenxuan Zhou and Muhao Chen were supported by the NSF Grants IIS 2105329 and ITE 2333736.

\bibliography{custom,anthology}

\begin{thebibliography}{72}
\providecommand{\natexlab}[1]{#1}
\providecommand{\url}[1]{\texttt{#1}}
\expandafter\ifx\csname urlstyle\endcsname\relax
  \providecommand{\doi}[1]{doi: #1}\else
  \providecommand{\doi}{doi: \begingroup \urlstyle{rm}\Url}\fi

\bibitem[Al-Rfou et~al.(2015)Al-Rfou, Kulkarni, Perozzi, and Skiena]{al2015polyglot}
Rami Al-Rfou, Vivek Kulkarni, Bryan Perozzi, and Steven Skiena.
\newblock Polyglot-ner: Massive multilingual named entity recognition.
\newblock In \emph{Proceedings of the 2015 SIAM International Conference on Data Mining}, pp.\  586--594. SIAM, 2015.

\bibitem[Arranz et~al.(2022)Arranz, Choukri, Cuadros, Garc{\'\i}a~Pablos, Gianola, Grouin, Herranz, Paroubek, and Zweigenbaum]{arranz-etal-2022-mapa}
Victoria Arranz, Khalid Choukri, Montse Cuadros, Aitor Garc{\'\i}a~Pablos, Lucie Gianola, Cyril Grouin, Manuel Herranz, Patrick Paroubek, and Pierre Zweigenbaum.
\newblock {MAPA} project: Ready-to-go open-source datasets and deep learning technology to remove identifying information from text documents.
\newblock In \emph{Proceedings of the Workshop on Ethical and Legal Issues in Human Language Technologies and Multilingual De-Identification of Sensitive Data In Language Resources within the 13th Language Resources and Evaluation Conference}, pp.\  64--72, Marseille, France, June 2022. European Language Resources Association.
\newblock URL \url{https://aclanthology.org/2022.legal-1.12}.

\bibitem[Au et~al.(2022)Au, Lampos, and Cox]{au2022ner}
Ting Wai~Terence Au, Vasileios Lampos, and Ingemar Cox.
\newblock {E}-{NER} {---} an annotated named entity recognition corpus of legal text.
\newblock In Nikolaos Aletras, Ilias Chalkidis, Leslie Barrett, C{\u{a}}t{\u{a}}lina Goan{\textcommabelow{t}}{\u{a}}, and Daniel Preo{\textcommabelow{t}}iuc-Pietro (eds.), \emph{Proceedings of the Natural Legal Language Processing Workshop 2022}, pp.\  246--255, Abu Dhabi, United Arab Emirates (Hybrid), December 2022. Association for Computational Linguistics.
\newblock \doi{10.18653/v1/2022.nllp-1.22}.
\newblock URL \url{https://aclanthology.org/2022.nllp-1.22}.

\bibitem[Chen et~al.(2022)Chen, Xu, Zhang, and Huang]{chen-etal-2022-crossroads}
Pei Chen, Haotian Xu, Cheng Zhang, and Ruihong Huang.
\newblock Crossroads, buildings and neighborhoods: A dataset for fine-grained location recognition.
\newblock In \emph{Proceedings of the 2022 Conference of the North American Chapter of the Association for Computational Linguistics: Human Language Technologies}, pp.\  3329--3339, Seattle, United States, July 2022. Association for Computational Linguistics.
\newblock \doi{10.18653/v1/2022.naacl-main.243}.
\newblock URL \url{https://aclanthology.org/2022.naacl-main.243}.

\bibitem[Chiang et~al.(2023)Chiang, Li, Lin, Sheng, Wu, Zhang, Zheng, Zhuang, Zhuang, Gonzalez, Stoica, and Xing]{vicuna-2023}
Wei-Lin Chiang, Zhuohan Li, Zi~Lin, Ying Sheng, Zhanghao Wu, Hao Zhang, Lianmin Zheng, Siyuan Zhuang, Yonghao Zhuang, Joseph~E. Gonzalez, Ion Stoica, and Eric~P. Xing.
\newblock Vicuna: An open-source chatbot impressing gpt-4 with 90\%* chatgpt quality, March 2023.
\newblock URL \url{https://vicuna.lmsys.org}.

\bibitem[Chung et~al.(2022)Chung, Hou, Longpre, Zoph, Tay, Fedus, Li, Wang, Dehghani, Brahma, et~al.]{chung2022scaling}
Hyung~Won Chung, Le~Hou, Shayne Longpre, Barret Zoph, Yi~Tay, William Fedus, Eric Li, Xuezhi Wang, Mostafa Dehghani, Siddhartha Brahma, et~al.
\newblock Scaling instruction-finetuned language models.
\newblock \emph{arXiv preprint arXiv:2210.11416}, 2022.

\bibitem[Collier \& Kim(2004)Collier and Kim]{collier2004introduction}
Nigel Collier and Jin-Dong Kim.
\newblock Introduction to the bio-entity recognition task at jnlpba.
\newblock In \emph{Proceedings of the International Joint Workshop on Natural Language Processing in Biomedicine and its Applications (NLPBA/BioNLP)}, pp.\  73--78, 2004.

\bibitem[Derczynski et~al.(2016)Derczynski, Bontcheva, and Roberts]{derczynski-etal-2016-broad}
Leon Derczynski, Kalina Bontcheva, and Ian Roberts.
\newblock Broad {T}witter corpus: A diverse named entity recognition resource.
\newblock In \emph{Proceedings of {COLING} 2016, the 26th International Conference on Computational Linguistics: Technical Papers}, pp.\  1169--1179, Osaka, Japan, December 2016. The COLING 2016 Organizing Committee.
\newblock URL \url{https://aclanthology.org/C16-1111}.

\bibitem[Ding et~al.(2021)Ding, Xu, Chen, Wang, Han, Xie, Zheng, and Liu]{ding-etal-2021-nerd}
Ning Ding, Guangwei Xu, Yulin Chen, Xiaobin Wang, Xu~Han, Pengjun Xie, Haitao Zheng, and Zhiyuan Liu.
\newblock Few-{NERD}: A few-shot named entity recognition dataset.
\newblock In \emph{Proceedings of the 59th Annual Meeting of the Association for Computational Linguistics and the 11th International Joint Conference on Natural Language Processing (Volume 1: Long Papers)}, pp.\  3198--3213, Online, August 2021. Association for Computational Linguistics.
\newblock \doi{10.18653/v1/2021.acl-long.248}.
\newblock URL \url{https://aclanthology.org/2021.acl-long.248}.

\bibitem[Do{\u{g}}an et~al.(2014)Do{\u{g}}an, Leaman, and Lu]{dougan2014ncbi}
Rezarta~Islamaj Do{\u{g}}an, Robert Leaman, and Zhiyong Lu.
\newblock Ncbi disease corpus: a resource for disease name recognition and concept normalization.
\newblock \emph{Journal of biomedical informatics}, 47:\penalty0 1--10, 2014.

\bibitem[Etzioni et~al.(2008)Etzioni, Banko, Soderland, and Weld]{etzioni2008open}
Oren Etzioni, Michele Banko, Stephen Soderland, and Daniel~S Weld.
\newblock Open information extraction from the web.
\newblock \emph{Communications of the ACM}, 51\penalty0 (12):\penalty0 68--74, 2008.

\bibitem[Friedrich et~al.(2020)Friedrich, Adel, Tomazic, Hingerl, Benteau, Maruscyk, and Lange]{friedrich2020sofcexp}
Annemarie Friedrich, Heike Adel, Federico Tomazic, Johannes Hingerl, Renou Benteau, Anika Maruscyk, and Lukas Lange.
\newblock The sofc-exp corpus and neural approaches to information extraction in the materials science domain, 2020.

\bibitem[Gao et~al.(2020)Gao, Biderman, Black, Golding, Hoppe, Foster, Phang, He, Thite, Nabeshima, et~al.]{gao2020pile}
Leo Gao, Stella Biderman, Sid Black, Laurence Golding, Travis Hoppe, Charles Foster, Jason Phang, Horace He, Anish Thite, Noa Nabeshima, et~al.
\newblock The pile: An 800gb dataset of diverse text for language modeling.
\newblock \emph{arXiv preprint arXiv:2101.00027}, 2020.

\bibitem[Grezes et~al.(2022)Grezes, Blanco-Cuaresma, Allen, and Ghosal]{grezes-etal-2022-overview}
Felix Grezes, Sergi Blanco-Cuaresma, Thomas Allen, and Tirthankar Ghosal.
\newblock Overview of the first shared task on detecting entities in the astrophysics literature ({DEAL}).
\newblock In \emph{Proceedings of the first Workshop on Information Extraction from Scientific Publications}, pp.\  1--7, Online, November 2022. Association for Computational Linguistics.
\newblock URL \url{https://aclanthology.org/2022.wiesp-1.1}.

\bibitem[Gu et~al.(2023)Gu, Zhang, Usuyama, Woldesenbet, Wong, Sanapathi, Wei, Valluri, Strandberg, Naumann, and Poon]{gu2023distilling}
Yu~Gu, Sheng Zhang, Naoto Usuyama, Yonas Woldesenbet, Cliff Wong, Praneeth Sanapathi, Mu~Wei, Naveen Valluri, Erika Strandberg, Tristan Naumann, and Hoifung Poon.
\newblock Distilling large language models for biomedical knowledge extraction: A case study on adverse drug events, 2023.

\bibitem[Guan et~al.(2023)Guan, Man, Chen, Yao, Hu, Zhu, Smith, Lim, and Yue]{guan2023findvehicle}
Runwei Guan, Ka~Lok Man, Feifan Chen, Shanliang Yao, Rongsheng Hu, Xiaohui Zhu, Jeremy Smith, Eng~Gee Lim, and Yutao Yue.
\newblock Findvehicle and vehiclefinder: A ner dataset for natural language-based vehicle retrieval and a keyword-based cross-modal vehicle retrieval system.
\newblock \emph{arXiv preprint arXiv:2304.10893}, 2023.

\bibitem[Gudibande et~al.(2023)Gudibande, Wallace, Snell, Geng, Liu, Abbeel, Levine, and Song]{gudibande2023false}
Arnav Gudibande, Eric Wallace, Charlie Snell, Xinyang Geng, Hao Liu, Pieter Abbeel, Sergey Levine, and Dawn Song.
\newblock The false promise of imitating proprietary llms, 2023.

\bibitem[Henry et~al.(2020)Henry, Buchan, Filannino, Stubbs, and Uzuner]{henry20202018}
Sam Henry, Kevin Buchan, Michele Filannino, Amber Stubbs, and Ozlem Uzuner.
\newblock 2018 n2c2 shared task on adverse drug events and medication extraction in electronic health records.
\newblock \emph{Journal of the American Medical Informatics Association}, 27\penalty0 (1):\penalty0 3--12, 2020.

\bibitem[Hinton et~al.(2015)Hinton, Vinyals, and Dean]{hinton2015distilling}
Geoffrey Hinton, Oriol Vinyals, and Jeff Dean.
\newblock Distilling the knowledge in a neural network.
\newblock \emph{arXiv preprint arXiv:1503.02531}, 2015.

\bibitem[Hsieh et~al.(2023)Hsieh, Li, Yeh, Nakhost, Fujii, Ratner, Krishna, Lee, and Pfister]{hsieh-etal-2023-distilling}
Cheng-Yu Hsieh, Chun-Liang Li, Chih-kuan Yeh, Hootan Nakhost, Yasuhisa Fujii, Alex Ratner, Ranjay Krishna, Chen-Yu Lee, and Tomas Pfister.
\newblock Distilling step-by-step! outperforming larger language models with less training data and smaller model sizes.
\newblock In \emph{Findings of the Association for Computational Linguistics: ACL 2023}, pp.\  8003--8017, Toronto, Canada, July 2023. Association for Computational Linguistics.
\newblock URL \url{https://aclanthology.org/2023.findings-acl.507}.

\bibitem[Hsieh et~al.(2022)Hsieh, Shih, and Ma]{hsieh-etal-2022-converting}
Yu-Ming Hsieh, Yueh-Yin Shih, and Wei-Yun Ma.
\newblock Converting the {S}inica {T}reebank of {M}andarin {C}hinese to {U}niversal {D}ependencies.
\newblock In \emph{Proceedings of the 16th Linguistic Annotation Workshop (LAW-XVI) within LREC2022}, pp.\  23--30, Marseille, France, June 2022. European Language Resources Association.
\newblock URL \url{https://aclanthology.org/2022.law-1.4}.

\bibitem[Jain et~al.(2020)Jain, van Zuylen, Hajishirzi, and Beltagy]{jain-etal-2020-scirex}
Sarthak Jain, Madeleine van Zuylen, Hannaneh Hajishirzi, and Iz~Beltagy.
\newblock {S}ci{REX}: {A} challenge dataset for document-level information extraction.
\newblock In \emph{Proceedings of the 58th Annual Meeting of the Association for Computational Linguistics}, pp.\  7506--7516, Online, July 2020. Association for Computational Linguistics.
\newblock \doi{10.18653/v1/2020.acl-main.670}.
\newblock URL \url{https://aclanthology.org/2020.acl-main.670}.

\bibitem[Jung et~al.(2023)Jung, West, Jiang, Brahman, Lu, Fisher, Sorensen, and Choi]{jung2023impossible}
Jaehun Jung, Peter West, Liwei Jiang, Faeze Brahman, Ximing Lu, Jillian Fisher, Taylor Sorensen, and Yejin Choi.
\newblock Impossible distillation: from low-quality model to high-quality dataset \& model for summarization and paraphrasing, 2023.

\bibitem[Kim et~al.(2003)Kim, Ohta, Tateisi, and Tsujii]{kim2003genia}
J-D Kim, Tomoko Ohta, Yuka Tateisi, and Jun’ichi Tsujii.
\newblock Genia corpus—a semantically annotated corpus for bio-textmining.
\newblock \emph{Bioinformatics}, 19\penalty0 (suppl\_1):\penalty0 i180--i182, 2003.

\bibitem[Krallinger et~al.(2015)Krallinger, Rabal, Leitner, Vazquez, Salgado, Lu, Leaman, Lu, Ji, Lowe, et~al.]{krallinger2015chemdner}
Martin Krallinger, Obdulia Rabal, Florian Leitner, Miguel Vazquez, David Salgado, Zhiyong Lu, Robert Leaman, Yanan Lu, Donghong Ji, Daniel~M Lowe, et~al.
\newblock The chemdner corpus of chemicals and drugs and its annotation principles.
\newblock \emph{Journal of cheminformatics}, 7\penalty0 (1):\penalty0 1--17, 2015.

\bibitem[Kulkarni et~al.(2018)Kulkarni, Xu, Ritter, and Machiraju]{kulkarni-etal-2018-annotated}
Chaitanya Kulkarni, Wei Xu, Alan Ritter, and Raghu Machiraju.
\newblock An annotated corpus for machine reading of instructions in wet lab protocols.
\newblock In \emph{Proceedings of the 2018 Conference of the North {A}merican Chapter of the Association for Computational Linguistics: Human Language Technologies, Volume 2 (Short Papers)}, pp.\  97--106, New Orleans, Louisiana, June 2018. Association for Computational Linguistics.
\newblock \doi{10.18653/v1/N18-2016}.
\newblock URL \url{https://aclanthology.org/N18-2016}.

\bibitem[Kumar \& Starly(2022)Kumar and Starly]{kumar2022fabner}
Aman Kumar and Binil Starly.
\newblock “fabner”: information extraction from manufacturing process science domain literature using named entity recognition.
\newblock \emph{Journal of Intelligent Manufacturing}, 33\penalty0 (8):\penalty0 2393--2407, 2022.

\bibitem[Li et~al.(2023)Li, Fang, Yang, Wang, Ye, Zhao, and Zhang]{li2023evaluating}
Bo~Li, Gexiang Fang, Yang Yang, Quansen Wang, Wei Ye, Wen Zhao, and Shikun Zhang.
\newblock Evaluating chatgpt's information extraction capabilities: An assessment of performance, explainability, calibration, and faithfulness.
\newblock \emph{arXiv preprint arXiv:2304.11633}, 2023.

\bibitem[Li et~al.(2016)Li, Sun, Johnson, Sciaky, Wei, Leaman, Davis, Mattingly, Wiegers, and Lu]{li2016biocreative}
Jiao Li, Yueping Sun, Robin~J Johnson, Daniela Sciaky, Chih-Hsuan Wei, Robert Leaman, Allan~Peter Davis, Carolyn~J Mattingly, Thomas~C Wiegers, and Zhiyong Lu.
\newblock Biocreative v cdr task corpus: a resource for chemical disease relation extraction.
\newblock \emph{Database}, 2016, 2016.

\bibitem[Liu et~al.(2013)Liu, Pasupat, Cyphers, and Glass]{liu2013asgard}
Jingjing Liu, Panupong Pasupat, Scott Cyphers, and Jim Glass.
\newblock Asgard: A portable architecture for multilingual dialogue systems.
\newblock In \emph{2013 IEEE International Conference on Acoustics, Speech and Signal Processing}, pp.\  8386--8390. IEEE, 2013.

\bibitem[Liu et~al.(2021)Liu, Xu, Yu, Dai, Ji, Cahyawijaya, Madotto, and Fung]{liu2021crossner}
Zihan Liu, Yan Xu, Tiezheng Yu, Wenliang Dai, Ziwei Ji, Samuel Cahyawijaya, Andrea Madotto, and Pascale Fung.
\newblock Crossner: Evaluating cross-domain named entity recognition.
\newblock In \emph{Proceedings of the AAAI Conference on Artificial Intelligence}, volume~35, pp.\  13452--13460, 2021.

\bibitem[Luan et~al.(2018)Luan, He, Ostendorf, and Hajishirzi]{luan-etal-2018-multi}
Yi~Luan, Luheng He, Mari Ostendorf, and Hannaneh Hajishirzi.
\newblock Multi-task identification of entities, relations, and coreference for scientific knowledge graph construction.
\newblock In \emph{Proceedings of the 2018 Conference on Empirical Methods in Natural Language Processing}, pp.\  3219--3232, Brussels, Belgium, October-November 2018. Association for Computational Linguistics.
\newblock \doi{10.18653/v1/D18-1360}.
\newblock URL \url{https://aclanthology.org/D18-1360}.

\bibitem[Luo et~al.(2022)Luo, Lai, Wei, Arighi, and Lu]{luo2022biored}
Ling Luo, Po-Ting Lai, Chih-Hsuan Wei, Cecilia~N Arighi, and Zhiyong Lu.
\newblock Biored: a rich biomedical relation extraction dataset.
\newblock \emph{Briefings in Bioinformatics}, 23\penalty0 (5):\penalty0 bbac282, 2022.

\bibitem[Mitchell et~al.(2005)Mitchell, Strassel, Huang, and Zakhary]{mitchell2005ace}
Alexis Mitchell, Stephanie Strassel, Shudong Huang, and Ramez Zakhary.
\newblock Ace 2004 multilingual training corpus.
\newblock \emph{Linguistic Data Consortium, Philadelphia}, 1:\penalty0 1--1, 2005.

\bibitem[Mowery et~al.(2014)Mowery, Velupillai, South, Christensen, Martinez, Kelly, Goeuriot, Elhadad, Pradhan, Savova, et~al.]{mowery2014task}
Danielle~L Mowery, Sumithra Velupillai, Brett~R South, Lee Christensen, David Martinez, Liadh Kelly, Lorraine Goeuriot, Noemie Elhadad, Sameer Pradhan, Guergana Savova, et~al.
\newblock Task 2: Share/clef ehealth evaluation lab 2014.
\newblock In \emph{Proceedings of CLEF 2014}, 2014.

\bibitem[Nye et~al.(2018)Nye, Li, Patel, Yang, Marshall, Nenkova, and Wallace]{nye2018corpus}
Benjamin Nye, Junyi~Jessy Li, Roma Patel, Yinfei Yang, Iain~J Marshall, Ani Nenkova, and Byron~C Wallace.
\newblock A corpus with multi-level annotations of patients, interventions and outcomes to support language processing for medical literature.
\newblock In \emph{Proceedings of the conference. Association for Computational Linguistics. Meeting}, volume 2018, pp.\  197. NIH Public Access, 2018.

\bibitem[OpenAI(2023)]{openai2023gpt4}
OpenAI.
\newblock Gpt-4 technical report, 2023.

\bibitem[Ouyang et~al.(2022)Ouyang, Wu, Jiang, Almeida, Wainwright, Mishkin, Zhang, Agarwal, Slama, Gray, et~al.]{ouyang2022training}
Long Ouyang, Jeffrey Wu, Xu~Jiang, Diogo Almeida, Carroll Wainwright, Pamela Mishkin, Chong Zhang, Sandhini Agarwal, Katarina Slama, Alex Gray, et~al.
\newblock Training language models to follow instructions with human feedback.
\newblock In \emph{Advances in Neural Information Processing Systems}, 2022.

\bibitem[Pan et~al.(2017)Pan, Zhang, May, Nothman, Knight, and Ji]{pan-etal-2017-cross}
Xiaoman Pan, Boliang Zhang, Jonathan May, Joel Nothman, Kevin Knight, and Heng Ji.
\newblock Cross-lingual name tagging and linking for 282 languages.
\newblock In \emph{Proceedings of the 55th Annual Meeting of the Association for Computational Linguistics (Volume 1: Long Papers)}, pp.\  1946--1958, Vancouver, Canada, July 2017. Association for Computational Linguistics.
\newblock \doi{10.18653/v1/P17-1178}.
\newblock URL \url{https://aclanthology.org/P17-1178}.

\bibitem[Peng et~al.(2023)Peng, Li, He, Galley, and Gao]{peng2023instruction}
Baolin Peng, Chunyuan Li, Pengcheng He, Michel Galley, and Jianfeng Gao.
\newblock Instruction tuning with gpt-4.
\newblock \emph{arXiv preprint arXiv:2304.03277}, 2023.

\bibitem[Perera et~al.(2020)Perera, Dehmer, and Emmert-Streib]{perera2020named}
Nadeesha Perera, Matthias Dehmer, and Frank Emmert-Streib.
\newblock Named entity recognition and relation detection for biomedical information extraction.
\newblock \emph{Frontiers in cell and developmental biology}, pp.\  673, 2020.

\bibitem[Pyysalo \& Ananiadou(2014)Pyysalo and Ananiadou]{pyysalo2014anatomical}
Sampo Pyysalo and Sophia Ananiadou.
\newblock Anatomical entity mention recognition at literature scale.
\newblock \emph{Bioinformatics}, 30\penalty0 (6):\penalty0 868--875, 2014.

\bibitem[Sanh et~al.(2021)Sanh, Webson, Raffel, Bach, Sutawika, Alyafeai, Chaffin, Stiegler, Scao, Raja, Dey, Bari, Xu, Thakker, Sharma, Szczechla, Kim, Chhablani, Nayak, Datta, Chang, Jiang, Wang, Manica, Shen, Yong, Pandey, Bawden, Wang, Neeraj, Rozen, Sharma, Santilli, Fevry, Fries, Teehan, Biderman, Gao, Bers, Wolf, and Rush]{sanh2021multitask}
Victor Sanh, Albert Webson, Colin Raffel, Stephen~H. Bach, Lintang Sutawika, Zaid Alyafeai, Antoine Chaffin, Arnaud Stiegler, Teven~Le Scao, Arun Raja, Manan Dey, M~Saiful Bari, Canwen Xu, Urmish Thakker, Shanya~Sharma Sharma, Eliza Szczechla, Taewoon Kim, Gunjan Chhablani, Nihal Nayak, Debajyoti Datta, Jonathan Chang, Mike Tian-Jian Jiang, Han Wang, Matteo Manica, Sheng Shen, Zheng~Xin Yong, Harshit Pandey, Rachel Bawden, Thomas Wang, Trishala Neeraj, Jos Rozen, Abheesht Sharma, Andrea Santilli, Thibault Fevry, Jason~Alan Fries, Ryan Teehan, Stella Biderman, Leo Gao, Tali Bers, Thomas Wolf, and Alexander~M. Rush.
\newblock Multitask prompted training enables zero-shot task generalization.
\newblock In \emph{International Conference on Learning Representations}, 2021.

\bibitem[Schindler et~al.(2021)Schindler, Bensmann, Dietze, and Kr{\"u}ger]{schindler2021somesci}
David Schindler, Felix Bensmann, Stefan Dietze, and Frank Kr{\"u}ger.
\newblock Somesci-a 5 star open data gold standard knowledge graph of software mentions in scientific articles.
\newblock In \emph{Proceedings of the 30th ACM International Conference on Information \& Knowledge Management}, pp.\  4574--4583, 2021.

\bibitem[Segura-Bedmar et~al.(2013)Segura-Bedmar, Mart{\'\i}nez~Fern{\'a}ndez, and Herrero~Zazo]{segura2013semeval}
Isabel Segura-Bedmar, Paloma Mart{\'\i}nez~Fern{\'a}ndez, and Mar{\'\i}a Herrero~Zazo.
\newblock Semeval-2013 task 9: Extraction of drug-drug interactions from biomedical texts (ddiextraction 2013).
\newblock Association for Computational Linguistics, 2013.

\bibitem[Shah et~al.(2023)Shah, Vithani, Gullapalli, and Chava]{shah2023finer}
Agam Shah, Ruchit Vithani, Abhinav Gullapalli, and Sudheer Chava.
\newblock Finer: Financial named entity recognition dataset and weak-supervision model.
\newblock \emph{arXiv preprint arXiv:2302.11157}, 2023.

\bibitem[Smith et~al.(2008)Smith, Tanabe, Kuo, Chung, Hsu, Lin, Klinger, Friedrich, Ganchev, Torii, et~al.]{smith2008overview}
Larry Smith, Lorraine~K Tanabe, Cheng-Ju Kuo, I~Chung, Chun-Nan Hsu, Yu-Shi Lin, Roman Klinger, Christoph~M Friedrich, Kuzman Ganchev, Manabu Torii, et~al.
\newblock Overview of biocreative ii gene mention recognition.
\newblock \emph{Genome biology}, 9\penalty0 (2):\penalty0 1--19, 2008.

\bibitem[Smith et~al.(2022)Smith, Fries, Hancock, and Bach]{smith2022language}
Ryan Smith, Jason~A Fries, Braden Hancock, and Stephen~H Bach.
\newblock Language models in the loop: Incorporating prompting into weak supervision.
\newblock \emph{arXiv preprint arXiv:2205.02318}, 2022.

\bibitem[Stubbs et~al.(2015)Stubbs, Kotfila, and Uzuner]{stubbs2015automated}
Amber Stubbs, Christopher Kotfila, and {\"O}zlem Uzuner.
\newblock Automated systems for the de-identification of longitudinal clinical narratives: Overview of 2014 i2b2/uthealth shared task track 1.
\newblock \emph{Journal of biomedical informatics}, 58:\penalty0 S11--S19, 2015.

\bibitem[Sun et~al.(2013)Sun, Rumshisky, and Uzuner]{sun2013evaluating}
Weiyi Sun, Anna Rumshisky, and Ozlem Uzuner.
\newblock Evaluating temporal relations in clinical text: 2012 i2b2 challenge.
\newblock \emph{Journal of the American Medical Informatics Association}, 20\penalty0 (5):\penalty0 806--813, 2013.

\bibitem[Sun et~al.(2023)Sun, Dong, Li, Wan, Wang, Zhang, Li, Cheng, Lyu, Wu, and Wang]{sun2023pushing}
Xiaofei Sun, Linfeng Dong, Xiaoya Li, Zhen Wan, Shuhe Wang, Tianwei Zhang, Jiwei Li, Fei Cheng, Lingjuan Lyu, Fei Wu, and Guoyin Wang.
\newblock Pushing the limits of chatgpt on nlp tasks, 2023.

\bibitem[Tabassum et~al.(2020)Tabassum, Maddela, Xu, and Ritter]{tabassum-etal-2020-code}
Jeniya Tabassum, Mounica Maddela, Wei Xu, and Alan Ritter.
\newblock Code and named entity recognition in {S}tack{O}verflow.
\newblock In \emph{Proceedings of the 58th Annual Meeting of the Association for Computational Linguistics}, pp.\  4913--4926, Online, July 2020. Association for Computational Linguistics.
\newblock \doi{10.18653/v1/2020.acl-main.443}.
\newblock URL \url{https://aclanthology.org/2020.acl-main.443}.

\bibitem[Taori et~al.(2023)Taori, Gulrajani, Zhang, Dubois, Li, Guestrin, Liang, and Hashimoto]{alpaca}
Rohan Taori, Ishaan Gulrajani, Tianyi Zhang, Yann Dubois, Xuechen Li, Carlos Guestrin, Percy Liang, and Tatsunori~B. Hashimoto.
\newblock Stanford alpaca: An instruction-following llama model.
\newblock \url{https://github.com/tatsu-lab/stanford_alpaca}, 2023.

\bibitem[Tedeschi \& Navigli(2022)Tedeschi and Navigli]{tedeschi-navigli-2022-multinerd}
Simone Tedeschi and Roberto Navigli.
\newblock {M}ulti{NERD}: A multilingual, multi-genre and fine-grained dataset for named entity recognition (and disambiguation).
\newblock In \emph{Findings of the Association for Computational Linguistics: NAACL 2022}, pp.\  801--812, Seattle, United States, July 2022. Association for Computational Linguistics.
\newblock \doi{10.18653/v1/2022.findings-naacl.60}.
\newblock URL \url{https://aclanthology.org/2022.findings-naacl.60}.

\bibitem[Tedeschi et~al.(2021)Tedeschi, Maiorca, Campolungo, Cecconi, and Navigli]{tedeschi-etal-2021-wikineural-combined}
Simone Tedeschi, Valentino Maiorca, Niccol{\`o} Campolungo, Francesco Cecconi, and Roberto Navigli.
\newblock {W}iki{NE}u{R}al: {C}ombined neural and knowledge-based silver data creation for multilingual {NER}.
\newblock In \emph{Findings of the Association for Computational Linguistics: EMNLP 2021}, pp.\  2521--2533, Punta Cana, Dominican Republic, November 2021. Association for Computational Linguistics.
\newblock \doi{10.18653/v1/2021.findings-emnlp.215}.
\newblock URL \url{https://aclanthology.org/2021.findings-emnlp.215}.

\bibitem[Touvron et~al.(2023{\natexlab{a}})Touvron, Lavril, Izacard, Martinet, Lachaux, Lacroix, Rozière, Goyal, Hambro, Azhar, Rodriguez, Joulin, Grave, and Lample]{touvron2023llama}
Hugo Touvron, Thibaut Lavril, Gautier Izacard, Xavier Martinet, Marie-Anne Lachaux, Timothée Lacroix, Baptiste Rozière, Naman Goyal, Eric Hambro, Faisal Azhar, Aurelien Rodriguez, Armand Joulin, Edouard Grave, and Guillaume Lample.
\newblock Llama: Open and efficient foundation language models, 2023{\natexlab{a}}.

\bibitem[Touvron et~al.(2023{\natexlab{b}})Touvron, Martin, Stone, Albert, Almahairi, Babaei, Bashlykov, Batra, Bhargava, Bhosale, Bikel, Blecher, Ferrer, Chen, Cucurull, Esiobu, Fernandes, Fu, Fu, Fuller, Gao, Goswami, Goyal, Hartshorn, Hosseini, Hou, Inan, Kardas, Kerkez, Khabsa, Kloumann, Korenev, Koura, Lachaux, Lavril, Lee, Liskovich, Lu, Mao, Martinet, Mihaylov, Mishra, Molybog, Nie, Poulton, Reizenstein, Rungta, Saladi, Schelten, Silva, Smith, Subramanian, Tan, Tang, Taylor, Williams, Kuan, Xu, Yan, Zarov, Zhang, Fan, Kambadur, Narang, Rodriguez, Stojnic, Edunov, and Scialom]{touvron2023llama2}
Hugo Touvron, Louis Martin, Kevin Stone, Peter Albert, Amjad Almahairi, Yasmine Babaei, Nikolay Bashlykov, Soumya Batra, Prajjwal Bhargava, Shruti Bhosale, Dan Bikel, Lukas Blecher, Cristian~Canton Ferrer, Moya Chen, Guillem Cucurull, David Esiobu, Jude Fernandes, Jeremy Fu, Wenyin Fu, Brian Fuller, Cynthia Gao, Vedanuj Goswami, Naman Goyal, Anthony Hartshorn, Saghar Hosseini, Rui Hou, Hakan Inan, Marcin Kardas, Viktor Kerkez, Madian Khabsa, Isabel Kloumann, Artem Korenev, Punit~Singh Koura, Marie-Anne Lachaux, Thibaut Lavril, Jenya Lee, Diana Liskovich, Yinghai Lu, Yuning Mao, Xavier Martinet, Todor Mihaylov, Pushkar Mishra, Igor Molybog, Yixin Nie, Andrew Poulton, Jeremy Reizenstein, Rashi Rungta, Kalyan Saladi, Alan Schelten, Ruan Silva, Eric~Michael Smith, Ranjan Subramanian, Xiaoqing~Ellen Tan, Binh Tang, Ross Taylor, Adina Williams, Jian~Xiang Kuan, Puxin Xu, Zheng Yan, Iliyan Zarov, Yuchen Zhang, Angela Fan, Melanie Kambadur, Sharan Narang, Aurelien Rodriguez, Robert Stojnic, Sergey Edunov, and Thomas
  Scialom.
\newblock Llama 2: Open foundation and fine-tuned chat models, 2023{\natexlab{b}}.

\bibitem[Ushio et~al.(2022)Ushio, Neves, Silva, Barbieri, and Camacho-Collados]{ushio-etal-2022-tweet}
Asahi Ushio, Leonardo Neves, Vitor Silva, Francesco. Barbieri, and Jose Camacho-Collados.
\newblock {N}amed {E}ntity {R}ecognition in {T}witter: {A} {D}ataset and {A}nalysis on {S}hort-{T}erm {T}emporal {S}hifts.
\newblock In \emph{The 2nd Conference of the Asia-Pacific Chapter of the Association for Computational Linguistics and the 12th International Joint Conference on Natural Language Processing}, Online, November 2022. Association for Computational Linguistics.

\bibitem[Uzuner et~al.(2007)Uzuner, Luo, and Szolovits]{uzuner2007evaluating}
{\"O}zlem Uzuner, Yuan Luo, and Peter Szolovits.
\newblock Evaluating the state-of-the-art in automatic de-identification.
\newblock \emph{Journal of the American Medical Informatics Association}, 14\penalty0 (5):\penalty0 550--563, 2007.

\bibitem[Uzuner et~al.(2011)Uzuner, South, Shen, and DuVall]{uzuner20112010}
{\"O}zlem Uzuner, Brett~R South, Shuying Shen, and Scott~L DuVall.
\newblock 2010 i2b2/va challenge on concepts, assertions, and relations in clinical text.
\newblock \emph{Journal of the American Medical Informatics Association}, 18\penalty0 (5):\penalty0 552--556, 2011.

\bibitem[Walker et~al.(2006)Walker, Strassel, Medero, and Maeda]{walker2006ace}
Christopher Walker, Stephanie Strassel, Julie Medero, and Kazuaki Maeda.
\newblock Ace 2005 multilingual training corpus.
\newblock \emph{Linguistic Data Consortium, Philadelphia}, 57:\penalty0 45, 2006.

\bibitem[Wang et~al.(2023{\natexlab{a}})Wang, Zhou, Zu, Xia, Chen, Zhang, Zheng, Ye, Zhang, Gui, Kang, Yang, Li, and Du]{wang2023instructuie}
Xiao Wang, Weikang Zhou, Can Zu, Han Xia, Tianze Chen, Yuansen Zhang, Rui Zheng, Junjie Ye, Qi~Zhang, Tao Gui, Jihua Kang, Jingsheng Yang, Siyuan Li, and Chunsai Du.
\newblock Instructuie: Multi-task instruction tuning for unified information extraction, 2023{\natexlab{a}}.

\bibitem[Wang et~al.(2022)Wang, Mishra, Alipoormolabashi, Kordi, Mirzaei, Naik, Ashok, Dhanasekaran, Arunkumar, Stap, Pathak, Karamanolakis, Lai, Purohit, Mondal, Anderson, Kuznia, Doshi, Pal, Patel, Moradshahi, Parmar, Purohit, Varshney, Kaza, Verma, Puri, Karia, Doshi, Sampat, Mishra, Reddy~A, Patro, Dixit, and Shen]{wang-etal-2022-super}
Yizhong Wang, Swaroop Mishra, Pegah Alipoormolabashi, Yeganeh Kordi, Amirreza Mirzaei, Atharva Naik, Arjun Ashok, Arut~Selvan Dhanasekaran, Anjana Arunkumar, David Stap, Eshaan Pathak, Giannis Karamanolakis, Haizhi Lai, Ishan Purohit, Ishani Mondal, Jacob Anderson, Kirby Kuznia, Krima Doshi, Kuntal~Kumar Pal, Maitreya Patel, Mehrad Moradshahi, Mihir Parmar, Mirali Purohit, Neeraj Varshney, Phani~Rohitha Kaza, Pulkit Verma, Ravsehaj~Singh Puri, Rushang Karia, Savan Doshi, Shailaja~Keyur Sampat, Siddhartha Mishra, Sujan Reddy~A, Sumanta Patro, Tanay Dixit, and Xudong Shen.
\newblock Super-{N}atural{I}nstructions: Generalization via declarative instructions on 1600+ {NLP} tasks.
\newblock In \emph{Proceedings of the 2022 Conference on Empirical Methods in Natural Language Processing}, pp.\  5085--5109, Abu Dhabi, United Arab Emirates, December 2022. Association for Computational Linguistics.
\newblock URL \url{https://aclanthology.org/2022.emnlp-main.340}.

\bibitem[Wang et~al.(2023{\natexlab{b}})Wang, Ivison, Dasigi, Hessel, Khot, Chandu, Wadden, MacMillan, Smith, Beltagy, and Hajishirzi]{wang2023far}
Yizhong Wang, Hamish Ivison, Pradeep Dasigi, Jack Hessel, Tushar Khot, Khyathi~Raghavi Chandu, David Wadden, Kelsey MacMillan, Noah~A. Smith, Iz~Beltagy, and Hannaneh Hajishirzi.
\newblock How far can camels go? exploring the state of instruction tuning on open resources, 2023{\natexlab{b}}.

\bibitem[Wang et~al.(2023{\natexlab{c}})Wang, Kordi, Mishra, Liu, Smith, Khashabi, and Hajishirzi]{wang2022self}
Yizhong Wang, Yeganeh Kordi, Swaroop Mishra, Alisa Liu, Noah~A. Smith, Daniel Khashabi, and Hannaneh Hajishirzi.
\newblock Self-instruct: Aligning language models with self-generated instructions.
\newblock In Anna Rogers, Jordan Boyd-Graber, and Naoaki Okazaki (eds.), \emph{Proceedings of the 61st Annual Meeting of the Association for Computational Linguistics (Volume 1: Long Papers)}, pp.\  13484--13508, Toronto, Canada, July 2023{\natexlab{c}}. Association for Computational Linguistics.
\newblock \doi{10.18653/v1/2023.acl-long.754}.
\newblock URL \url{https://aclanthology.org/2023.acl-long.754}.

\bibitem[Wang et~al.(2019)Wang, Shang, Liu, Lu, Liu, and Han]{wang2019crossweigh}
Zihan Wang, Jingbo Shang, Liyuan Liu, Lihao Lu, Jiacheng Liu, and Jiawei Han.
\newblock Crossweigh: Training named entity tagger from imperfect annotations.
\newblock In \emph{Proceedings of the 2019 Conference on Empirical Methods in Natural Language Processing and the 9th International Joint Conference on Natural Language Processing (EMNLP-IJCNLP)}, pp.\  5157--5166, 2019.

\bibitem[Wei et~al.(2021)Wei, Bosma, Zhao, Guu, Yu, Lester, Du, Dai, and Le]{wei2022finetuned}
Jason Wei, Maarten Bosma, Vincent Zhao, Kelvin Guu, Adams~Wei Yu, Brian Lester, Nan Du, Andrew~M Dai, and Quoc~V Le.
\newblock Finetuned language models are zero-shot learners.
\newblock In \emph{International Conference on Learning Representations}, 2021.

\bibitem[Wei et~al.(2023)Wei, Cui, Cheng, Wang, Zhang, Huang, Xie, Xu, Chen, Zhang, et~al.]{wei2023zero}
Xiang Wei, Xingyu Cui, Ning Cheng, Xiaobin Wang, Xin Zhang, Shen Huang, Pengjun Xie, Jinan Xu, Yufeng Chen, Meishan Zhang, et~al.
\newblock Zero-shot information extraction via chatting with chatgpt.
\newblock \emph{arXiv preprint arXiv:2302.10205}, 2023.

\bibitem[Weischedel et~al.(2013)Weischedel, Palmer, Marcus, Hovy, Pradhan, Ramshaw, Xue, Taylor, Kaufman, Franchini, et~al.]{weischedel2013ontonotes}
Ralph Weischedel, Martha Palmer, Mitchell Marcus, Eduard Hovy, Sameer Pradhan, Lance Ramshaw, Nianwen Xue, Ann Taylor, Jeff Kaufman, Michelle Franchini, et~al.
\newblock Ontonotes release 5.0 ldc2013t19.
\newblock \emph{Linguistic Data Consortium, Philadelphia, PA}, 23:\penalty0 170, 2013.

\bibitem[Wr{\'o}blewska et~al.(2022)Wr{\'o}blewska, Kaliska, Paw{\l}owski, Wi{\'s}niewski, Sosnowski, and {\L}awrynowicz]{wroblewska2022tasteset}
Ania Wr{\'o}blewska, Agnieszka Kaliska, Maciej Paw{\l}owski, Dawid Wi{\'s}niewski, Witold Sosnowski, and Agnieszka {\L}awrynowicz.
\newblock Tasteset--recipe dataset and food entities recognition benchmark.
\newblock \emph{arXiv preprint arXiv:2204.07775}, 2022.

\bibitem[Wu et~al.(2017)Wu, Jiang, Xu, Zhi, and Xu]{wu2017clinical}
Yonghui Wu, Min Jiang, Jun Xu, Degui Zhi, and Hua Xu.
\newblock Clinical named entity recognition using deep learning models.
\newblock In \emph{AMIA annual symposium proceedings}, volume 2017, pp.\  1812. American Medical Informatics Association, 2017.

\bibitem[Ye et~al.(2023)Ye, Chen, Xu, Zu, Shao, Liu, Cui, Zhou, Gong, Shen, et~al.]{ye2023comprehensive}
Junjie Ye, Xuanting Chen, Nuo Xu, Can Zu, Zekai Shao, Shichun Liu, Yuhan Cui, Zeyang Zhou, Chao Gong, Yang Shen, et~al.
\newblock A comprehensive capability analysis of gpt-3 and gpt-3.5 series models.
\newblock \emph{arXiv preprint arXiv:2303.10420}, 2023.

\end{thebibliography}
\bibliographystyle{iclr2024_conference}

\newpage
\appendix
\section{Appendix}

\subsection{Case Study}

\paragraph{Sensitivity to entity type paraphrasing.}
One type of entity can be expressed in multiple different ways.
In this scenario, it is essential for our model to give consistent predictions given entity types with similar meanings.
An example of sensitivity analysis is present in Fig.~\ref{fig:sensitivity}.
We observe that \shortname-7B-type 
sometimes fails to recognize entities with similar semantic meanings.
On the other hand, \shortname-7B-definition, despite performing worse on our Universal NER benchmark, exhibits robustness to entity type paraphrasing.
It demonstrates that although using definitions may result in lower performance on standard NER benchmarks, it could yield improved performance for less populous entity types.

\begin{figure}[!ht]
    \begin{AIboxSmall}{\footnotesize Sensitivity Analysis of Entity Types}
    \footnotesize
    Text: I'm visiting Los Angeles next week.\\

    \texttt{User:} What describes city in the text?\\
    \texttt{\shortname-7B-type}: \colorbox{GREEN}{["Los Angeles"]} \\
    \texttt{\shortname-7B-definition}: \colorbox{GREEN}{["Los Angeles"]} \\

    \texttt{User:} What describes place in the text?\\
    \texttt{\shortname-7B-type}: \colorbox{RED}{[]} \\
    \texttt{\shortname-7B-definition}: \colorbox{GREEN}{["Los Angeles"]} \\

    \texttt{User:} What describes metropolis in the text?\\
    \texttt{\shortname-7B-type}: \colorbox{RED}{[]} \\
    \texttt{\shortname-7B-definition}: \colorbox{GREEN}{["Los Angeles"]} \\

    \texttt{User:} What describes urban area in the text?\\
    \texttt{\shortname-7B-type}: \colorbox{RED}{[]} \\
    \texttt{\shortname-7B-definition}: \colorbox{GREEN}{["Los Angeles"]} \\

    \texttt{User:} What describes human settlement in the text? \\
    \texttt{\shortname-7B-type}: \colorbox{RED}{[]} \\
    \texttt{\shortname-7B-definition}: \colorbox{GREEN}{["Los Angeles"]}
    \end{AIboxSmall}
    \caption{Sensitivity of \shortname-7B on different entity types. \colorbox{GREEN}{Green} and \colorbox{RED}{red} mark correct and wrong predictions, respectively.}
    \label{fig:sensitivity}
\end{figure}

\paragraph{Recognition of diverse entity types}

The code snippet provided is for text generation using the Huggingface library:
\definecolor{solarizedBase03}{RGB}{0, 43, 54}
\definecolor{solarizedBase02}{RGB}{7, 54, 66}
\definecolor{solarizedBase01}{RGB}{88, 110, 117}
\definecolor{solarizedBase00}{RGB}{101, 123, 131}
\definecolor{solarizedBase0}{RGB}{131, 148, 150}
\definecolor{solarizedBase1}{RGB}{147, 161, 161}
\definecolor{solarizedBase2}{RGB}{238, 232, 213}
\definecolor{solarizedBase3}{RGB}{253, 246, 227}
\definecolor{solarizedYellow}{RGB}{181, 137, 0}
\definecolor{solarizedOrange}{RGB}{203, 75, 22}
\definecolor{solarizedRed}{RGB}{220, 50, 47}
\definecolor{solarizedMagenta}{RGB}{211, 54, 130}
\definecolor{solarizedViolet}{RGB}{108, 113, 196}
\definecolor{solarizedBlue}{RGB}{38, 139, 210}
\definecolor{solarizedCyan}{RGB}{42, 161, 152}
\definecolor{solarizedGreen}{RGB}{133, 153, 0}

\lstset{
  language=Python,
  basicstyle=\linespread{1.2}\scriptsize\fontfamily{qhv}\selectfont\color{solarizedBase03},
  keywordstyle=\color{solarizedGreen},
  stringstyle=\color{solarizedCyan},
  commentstyle=\color{solarizedBase01},
  numberstyle=\scriptsize\color{solarizedBase01},
  numbersep=5pt,
  breaklines=true,
  frame=single,
  rulecolor=\color{black},
  linewidth=0.8\textwidth,
  postbreak=\mbox{\textcolor{solarizedRed}{$\hookrightarrow$}\space},
}

\begin{figure}[!ht]
\begin{center}
\vspace{-10pt}
\begin{tabular}{c}
\begin{lstlisting}
from transformers import GPT2Tokenizer, GPT2LMHeadModel

tokenizer = GPT2Tokenizer.from_pretrained('gpt2')
model = GPT2LMHeadModel.from_pretrained('gpt2')

input_text = "This is a simple program"
input_ids = tokenizer.encode(input_text, return_tensors='pt')
output = model.generate(input_ids, max_new_tokens=50)
generated_text = tokenizer.decode(output[0], skip_special_tokens=True)
\end{lstlisting}
\end{tabular}
\end{center}
    \caption{A code snippet to assess \shortname's ability to extract diverse types of entities within the code.}
    \label{fig:code-ex}
\end{figure}

In the ensuing conversations with \shortname, we present the queries and corresponding results:

\begin{tcolorbox}[boxsep=0pt]
\small
\texttt{User:} What describes library in the text? \\
\texttt{Assistant:} ["transformers"] \\
\texttt{User:} What describes local variable in the text? \\
\texttt{Assistant:} ["input\_text", "input\_ids", "output", "generated\_text"] \\
\texttt{User:} What describes global variable in the text? \\
\texttt{Assistant:} [] \\
\texttt{User:} What describes method in the text? \\
\texttt{Assistant:} ["encode", "generate", "decode"] \\
\texttt{User:} What describes class in the text? \\
\texttt{Assistant:}["GPT2Tokenizer","GPT2LMHeadModel"]
\end{tcolorbox}

\smallskip
Despite minor mistakes (e.g., from\_pretrained is not identified as a method), this case study effectively demonstrates our model's capacity to capture entities of various types.

\section{Full Evaluation Results}
\begin{figure}[!ht]
\centering
\includegraphics[width=0.43\textwidth]{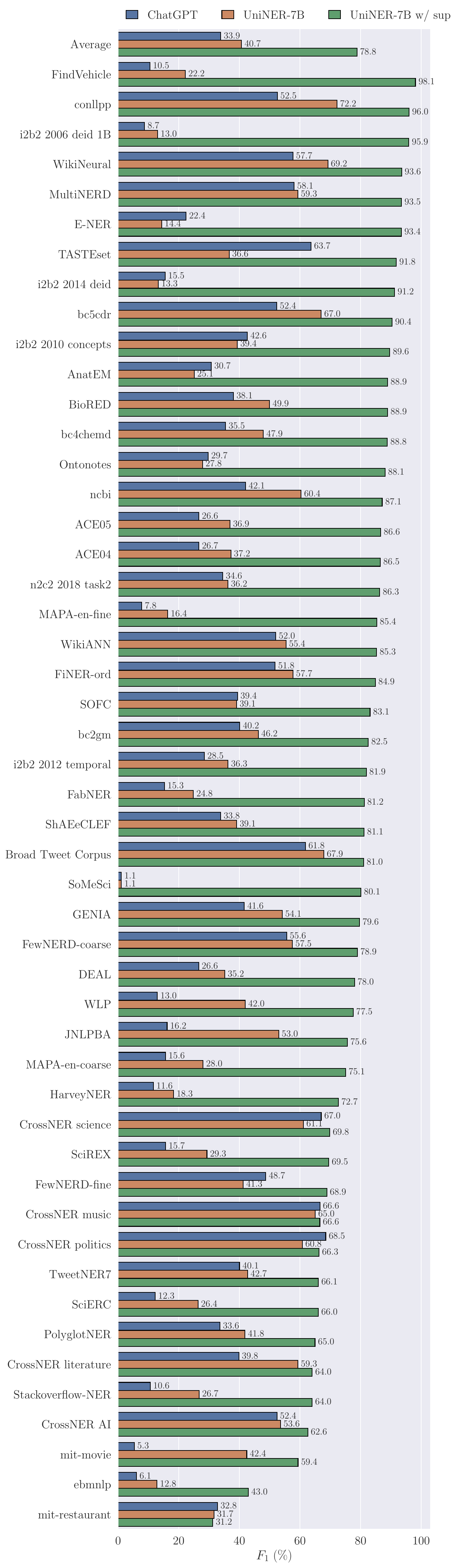}
\caption{Full evaluation results of ChatGPT, \shortname-7B, and \shortname{}-7B w/ sup (joint training on both supervised and Pile-type data, MIT and CrossNER data are excluded in training).}
\label{fig:add-supervised-bar}
\end{figure}

Full results on ChatGPT, \shortname-7B-type, and \shortname-7B-sup+type are shown in Fig.~\ref{fig:add-supervised-bar}.

\section{Data Statistics}
We show the full dataset statistics in Universal NER in Tab.~\ref{tab:full_data_statistics}, including the number of instances in train/dev/test data, number of entity types, average number of tokens in input text, and the average number of entities in each instance.

\begin{table*}[!ht]
\centering
\tablestyle{2pt}{1.2}
\begin{tabular}{c|lcccccc}
\shline
& &  &  & & & Avg.& Avg. \\
Domain & Dataset & \# train & \# dev & \# test & \# types& tokens& entities  \\ \shline
\multirow{16}{*}{General}& ACE04~\citep{mitchell2005ace}& 6202& 745& 812& 7& 37& 4.5 \\
& ACE05~\citep{walker2006ace}& 7299& 971& 1060& 7& 21& 2.8 \\
& conllpp~\citep{wang2019crossweigh}& 14041& 3250& 3453& 3& 25& 1.9 \\
& CrossNER AI~\citep{liu2021crossner}& 100& 350& 431& 13& 52& 5.3 \\
& CrossNER literature~\citep{liu2021crossner}& 100& 400& 416& 11& 54& 5.4 \\
& CrossNER music~\citep{liu2021crossner}& 100& 380& 465& 12& 57& 6.5 \\
& CrossNER politics~\citep{liu2021crossner}& 199& 540& 650& 8& 61& 6.5 \\
& CrossNER science~\citep{liu2021crossner}& 200& 450& 543& 16& 54& 5.4 \\
& FewNERD-coarse~\citep{ding-etal-2021-nerd}& 131767& 18824& 37648& 7& 35& 2.6 \\
& FewNERD-fine~\citep{ding-etal-2021-nerd}& 131767& 18824& 37648& 59& 35& 2.6 \\
& MultiNERD~\citep{tedeschi-navigli-2022-multinerd}& 134144& 10000& 10000& 16& 28& 1.6 \\
& Ontonotes~\citep{weischedel2013ontonotes}& 59924& 8528& 8262& 18& 18& 0.9 \\
& PolyglotNER~\citep{al2015polyglot}& 393982& 10000& 10000& 3& 34& 1.0 \\
& TASTEset~\citep{wroblewska2022tasteset}& 556& 69& 71& 9& 62& 19.1 \\
& WikiANN en~\citep{pan-etal-2017-cross}& 20000& 10000& 10000& 3& 15& 1.4 \\
& WikiNeural~\citep{tedeschi-etal-2021-wikineural-combined}& 92720& 11590& 11597& 3& 33& 1.4 \\
\shline
\multirow{8}{*}{Biomed}& AnatEM~\citep{pyysalo2014anatomical}& 5861& 2118& 3830& 1& 37& 0.7 \\
& BioRED~\citep{luo2022biored} & 4373& 1131& 1106& 6& 46& 3.0 \\
& GENIA~\citep{kim2003genia} & 15023& 1669& 1854& 5& 43& 3.5 \\
& JNLPBA~\citep{collier2004introduction} & 18608& 1940& 4261& 5& 39& 2.8 \\
& bc2gm~\citep{smith2008overview} & 12500& 2500& 5000& 1& 36& 0.4 \\
& bc4chemd~\citep{krallinger2015chemdner} & 30682& 30639& 26364& 1& 45& 0.9 \\
& bc5cdr~\citep{li2016biocreative} & 4560& 4581& 4797& 2& 41& 2.2 \\
& ncbi~\citep{dougan2014ncbi} & 5432& 923& 940& 1& 39& 1.0 \\
\shline
\multirow{6}{*}{Clinics}& ebmnlp~\citep{nye2018corpus} & 40713& 10608& 2076& 3& 43& 1.7 \\
& i2b2 2006 deid 1B~\citep{uzuner2007evaluating} & 34958& 14983& 18095& 8& 16& 0.3 \\
& i2b2 2010 concepts~\citep{uzuner20112010} & 14553& 1762& 27625& 3& 18& 1.0 \\
& i2b2 2012 temporal~\citep{sun2013evaluating} & 6235& 787& 5282& 6& 22& 2.3 \\
& i2b2 2014 deid~\citep{stubbs2015automated} & 46272& 4610& 32587& 23& 21& 0.4 \\
& n2c2 2018 task2~\citep{henry20202018} & 84351& 9252& 60228& 9& 14& 0.6 \\
& ShAEeCLEF~\citep{mowery2014task} & 12494& 2459& 14143& 1& 13& 0.3 \\
\shline
\multirow{6}{*}{STEM}& DEAL~\citep{grezes-etal-2022-overview} & 26906& 20800& 36665& 30& 35& 1.4 \\
& FabNER~\citep{kumar2022fabner} & 9435& 2182& 2064& 12& 36& 5.1 \\
& SOFC~\citep{friedrich2020sofcexp} & 568& 135& 173& 3& 68& 5.3 \\
& SciERC~\citep{luan-etal-2018-multi} & 350& 50& 100& 4& 163& 16.0 \\
& SciREX~\citep{jain-etal-2020-scirex} & 71511& 15182& 16599& 4& 29& 1.4 \\
& SoMeSci~\citep{schindler2021somesci} & 31055& 159& 16427& 14& 41& 2.4 \\
& WLP~\citep{kulkarni-etal-2018-annotated} & 8177& 2717& 2726& 16& 25& 4.5 \\
\shline
Programming & Stackoverflow-NER~\citep{tabassum-etal-2020-code}& 9263& 2936& 3108& 25& 19& 1.2 \\
\shline
\multirow{5}{*}{Social media}& HarveyNER~\citep{chen-etal-2022-crossroads} & 3967& 1301& 1303& 4& 48& 0.4 \\
& Broad Tweet Corpus~\citep{derczynski-etal-2016-broad} & 5334& 2001& 2000& 3& 28& 0.5 \\
& TweetNER7~\citep{ushio-etal-2022-tweet} & 7111& 886& 576& 7& 52& 3.1 \\
& mit-movie~\citep{liu2013asgard} & 9774& 2442& 2442& 12& 13& 1.8 \\
& mit-restaurant~\citep{liu2013asgard} & 7659& 1520& 1520& 8& 13& 2.2 \\
\shline
\multirow{3}{*}{Law}& E-NER~\citep{au2022ner} & 8072& 1009& 1010& 6& 55& 0.8 \\
& MAPA-coarse~\citep{arranz-etal-2022-mapa} & 893& 98& 408& 5& 56& 0.9 \\
& MAPA-fine~\citep{arranz-etal-2022-mapa} & 893& 98& 408& 17& 56& 1.3 \\
\shline
Finance & FiNER-ord~\citep{shah2023finer} & 3262& 403& 1075& 3& 34& 1.1 \\
\shline
Transportation & FindVehicle~\citep{guan2023findvehicle} & 21565& 20777& 20777& 21& 33& 5.5 \\
\shline
\end{tabular}
\caption{Statistics of datasets in our benchmark.}
\label{tab:full_data_statistics}
\end{table*}

\end{document}